\documentclass[letterpaper]{article}
\usepackage{aaai2026} 
\usepackage{times} 
\usepackage{helvet} 
\usepackage{courier} 
\usepackage[hyphens]{url} 
\usepackage{graphicx}
\usepackage{natbib}
\usepackage{caption}
\usepackage{algorithm}
\usepackage{algorithmic}

\usepackage{amsmath}
\usepackage{amssymb}
\usepackage{mathtools}
\usepackage{amsthm}
\usepackage{multirow}
\usepackage[table]{xcolor}
\usepackage{booktabs}

\definecolor{deeppink}{rgb}{0.62, 1.00, 0.84} 
\definecolor{deep_blue}{rgb}{0.00, 0.45, 0.72} 

\newcommand{\LS}[1]{\left({#1}\right)}
\newcommand{\LM}[1]{\left\{{#1}\right\}}
\newcommand{\LL}[1]{\left[{#1}\right]}

\newcommand{\C}[1]{\mathcal{#1}}
\newcommand{\R}{\mathbb{R}}

\newcommand{\B}[1]{\boldsymbol{#1}}

\newcommand{\OP}[1]{\mathtt{#1}}

\usepackage{subcaption}

\theoremstyle{plain}

\theoremstyle{definition}

\theoremstyle{remark}

\usepackage{newfloat}
\usepackage{listings}
\DeclareCaptionStyle{ruled}{labelfont=normalfont,labelsep=colon,strut=off}
\floatstyle{ruled}
\newfloat{listing}{tb}{lst}{}
\floatname{listing}{Listing}

\nocopyright 

\title{REXO: Indoor Multi-View Radar Object Detection via \\3D Bounding Box Diffusion}
\author{
    Ryoma Yataka\textsuperscript{\rm 1,2}\thanks{The work was done as a visiting scientist of MERL from ITC.}, Pu (Perry) Wang\textsuperscript{\rm 2}, Petros Boufounos\textsuperscript{\rm 2}, Ryuhei Takahashi\textsuperscript{\rm 1}\\
}
\affiliations{
    \textsuperscript{\rm 1}\textit{Information Technology R\&D Center (ITC), Mitsubishi Electric Corporation (MELCO)}, Kanagawa 247-8501, Japan\\
    \textsuperscript{\rm 2}\textit{Mitsubishi Electric Research Laboratories (MERL)}, Cambridge, MA 02139, USA
}

\begin{document}

\maketitle

\begin{abstract}
Multi-view indoor radar perception has drawn attention due to its cost-effectiveness and low privacy risks. Existing methods often rely on {implicit} cross-view radar feature association, such as proposal pairing in RFMask or query-to-feature cross-attention in RETR, which can lead to ambiguous feature matches and degraded detection in complex indoor scenes. To address these limitations, we propose \textbf{REXO} (multi-view Radar object dEtection with 3D bounding boX diffusiOn), which lifts the 2D bounding box (BBox) diffusion process of DiffusionDet into the 3D radar space. REXO utilizes these noisy 3D BBoxes to guide an {explicit} cross-view radar feature association, enhancing  the cross-view radar-conditioned denoising process. By accounting for prior knowledge that the person is in contact with the ground, REXO reduces the number of diffusion parameters by determining them from this prior. Evaluated on two open indoor radar datasets, our approach surpasses state-of-the-art methods by a margin of $+4.22$ AP on the HIBER dataset and $+11.02$ AP on the MMVR dataset. The REXO implementation is available at \mbox{\url{https://github.com/merlresearch/radar-bbox-diffusion}}.
\end{abstract}

\section{Introduction}
\label{sec:intro}

\begin{figure}[t]
    \centering
    \includegraphics[width=0.9\linewidth]{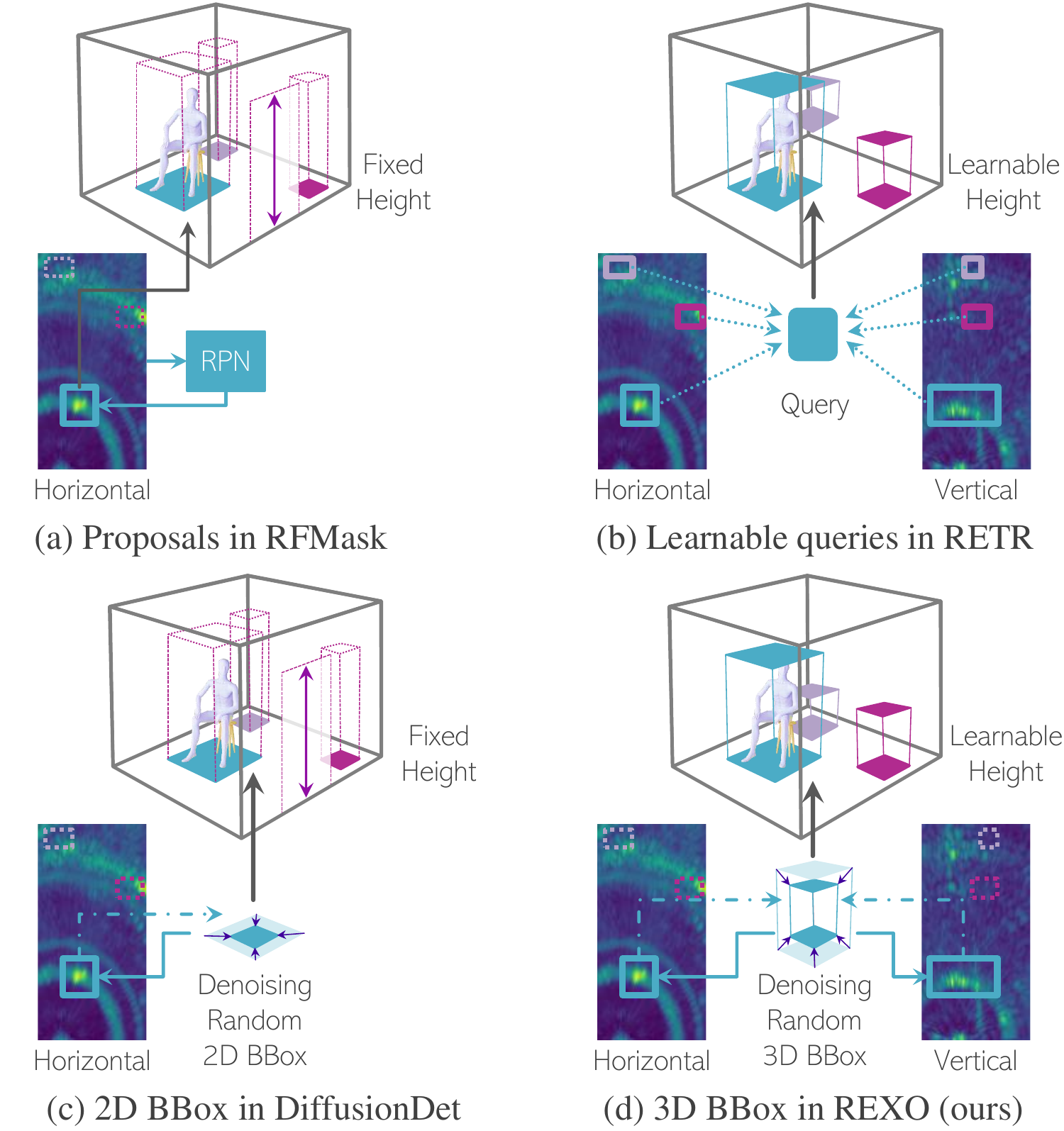}
    \vspace{-3mm}
    \caption{
        (a) RFMask~\cite{Wu2023_RFMask} generates horizontal-view proposals with fixed-height vertical boxes;
        (b) RETR~\cite{Yataka2024_retr} implicitly links queries to cross-view features via decoder cross-attention;
        (c) DiffusionDet~\cite{Chen2023_diffusiondet} adapted to horizontal radar allows 2D denoising but needs extra pairing with fixed-height vertical boxes;
        (d) REXO (\textbf{ours}) performs diffusion directly in 3D radar space for simple, explicit cross-view association.
    }
    \label{fig:comparison_obj_detection}
    \vspace{-6mm}
\end{figure}

Radar perception has received increasing attention due to its robustness in low-light, adversarial weather, and hazardous conditions (e.g., smoke)~\cite{Peak2023_KRadar, Yao2024, Lu20a, Sun2020_MIMORadar_AutonomousDriging, Pandharipande2023_SensingMachineLearning, Skog2024_humandetection4dradar}. Depending on the application, operational specifications, and downstream tasks,  radar data can be represented in different forms such as sparse detection points \cite{Mingmin2017_RFSleep, Sengupta2020_mmPose, Yang2023_mmFi}, reflection heatmaps \cite{Fadel2015_RFCapture, Zhao2018_RFPose, Wu2023_RFMask}, Doppler signatures, and raw analog-to-digital converter (ADC) data, each with unique characteristics and feature granularity. 

For indoor radar perception, single‑view and multi‑view heatmaps that combine horizontal (depth‑horizontal) and vertical (depth‑vertical) projections enable object detection, pose estimation, and segmentation on a 2D image plane~\cite{Zhao2018_RFPose, Lee2023_HuPR, Wu2023_RFMask, Perry2024_MMVR}. RF‑Pose~\cite{Zhao2018_RFPose, Zhao2018_RFPose3D} first fused both views with a convolutional autoencoder to regress 2D human keypoints. RFMask~\cite{Wu2023_RFMask} grafted Faster R‑CNN~\cite{Ren2017_FasterRCNN} onto horizontal heatmaps: its region‑proposal network produces horizontal candidates that are paired with fixed‑height vertical windows to avoid exhaustive cross‑view association (Fig.~\ref{fig:comparison_obj_detection} (a)). More recently, the radar detection transformer (RETR)~\cite{Yataka2024_retr} adopted the DETR~\cite{Carion2020_detr}. Decoder queries simultaneously attend to both views through cross‑attention (Fig.~\ref{fig:comparison_obj_detection} (b)) and are directly regressed to 3D bounding boxes (BBoxes) that are classified as person or background.

On the other hand, image-based object detection has been redefined as a generative denoising process, where random noisy 2D BBoxes are iteratively refined through a diffusion denoising process to yield final clean BBox predictions \cite{Chen2023_diffusiondet}. Referred to as DiffusionDet, it decouples training and inference, and generally surpasses query‑based detectors. When ported to horizontal radar heatmaps (Fig.~\ref{fig:comparison_obj_detection} (c)), it denoises 2D boxes but still requires the fixed‑height vertical pairing used by RFMask.

We therefore \emph{lift} the diffusion procedure from a 2D plane (image or horizontal radar view) in DiffusionDet to the full 3D radar space, as illustrated in Fig.~\ref{fig:comparison_obj_detection} (d). This simple lifting facilitates cross-view radar feature association and radar-conditioned BBox denoising, while enabling the integration of geometry-aware loss functions and prior constraints on the 3D BBox. Consequently, we introduce the proposed framework as \textbf{Radar object dEtection with 3D bounding boX diffusiOn (REXO)} with the following contributions:
\begin{enumerate}
    \item \textbf{2D‑to‑3D Lifting with Explicit Cross‑View Association}: At each diffusion timestep, noisy 3D BBoxes are projected onto every radar view, and RoI‑aligned crops supply view‑specific features. This BBox‑guided association grows \emph{linearly} with the number of views, whereas proposal‑ or query‑based schemes grow quadratically.
    \item \textbf{Cross-View Radar-Conditioned BBox Detection}: While the cross-view feature association is simplified due to the 2D-to-3D lifting, the denoising process may be more challenging. In turn, the associated radar features are used as conditioning to alleviate the more challenging 3D BBox denoising. To the best of our knowledge, REXO is the first diffusion model in the radar perception field conditioned on multi-view radar.
    \item \textbf{Ground-Level Constraint}: By using prior knowledge that the person is in contact with the ground, the parameters of the 3D BBox are reduced. Based on this, each noise-free 3D BBox preserves geometric constraints in the image plane to be transformed.
\end{enumerate}
We demonstrate the effectiveness of our contributions through evaluations on two open radar datasets.

\section{Related Work}
\label{sec:related_work}

\paragraph{Radar-based Object Detection:} Learning-based methods have advanced radar detection over traditional model-based approaches~\cite{Kay98}, benefiting from open large-scale radar point cloud datasets like nuScenes~\cite{caesar2020nuscenes}, Oxford RobotCar~\cite{Oxford20}, and RADIATE~\cite{Sheeny2021_RADIATE}.  Image-based and point/voxel-based backbones \cite{He2016_ResNet, PillarNet} extract semantic features from radar detection points, generate region proposals, and localize objects. High-resolution heatmaps (e.g., K-Radar~\cite{Peak2023_KRadar}, HIBER~\cite{Wu2023_RFMask}, MMVR~\cite{Perry2024_MMVR}) and raw ADC data~\cite{ADCNet} have also been leveraged by previously mentioned RF-Pose \cite{Zhao2018_RFPose}, RFMask \cite{Wu2023_RFMask}, and RETR \cite{Yataka2024_retr}. CubeLearn \cite{CubeLearn} replaces Fourier transforms with learnable modules for an end-to-end radar pipeline, while RAMP-CNN \cite{RAMPCNN} enhances range-angle feature extraction via Doppler cues. More recently, diffusion models have been explored for radar applications~\cite{Zhang2024_DiffusionRadar,Luan2024_DiffPCSuperResol,chi2024_rfdiffusionradiosignalgeneration,Fan2024_RFVision,Wu2024_DiffRadar}. Most efforts, e.g., Radar-Diffusion \cite{Zhang2024_DiffusionRadar, Luan2024_DiffPCSuperResol} and DiffRadar \cite{Wu2024_DiffRadar}, focus on reconstructing LiDAR-like point clouds from low-resolution radar data, while mmDiff \cite{Fan2024_RFVision} estimates and refines pose keypoints from sparse radar points via diffusion process. 

\paragraph{Diffusion-based Object Detection:} 
Diffusion models \cite{Song2021_DDIM,Song2019_ScoreBasedModel,Rombach2022_StableDiff,Song2023_ConsistencyModel} have shown impressive results in tasks such as image and video generation~\cite{Ho2022_VideoGen,Blattmann2023_AlignYourLatent} and multi-view synthesis~\cite{Chen2023_3DView,Yu2023_ICCV_LongTerm3DView}. For perception tasks, DiffusionDet~\cite{Chen2023_diffusiondet} first reformulates object detection as a generative denoising process and proposes to model the 2D BBoxes as random parameters in the diffusion process. Diffusion-SS3D~\cite{Ho2023_DiffSS3D} proposes a diffusion-based detector to enhance the quality of pseudo-labels in semi-supervised 3D object detection by integrating it into a teacher-student framework. CLIFF~\cite{Li2024_cliff} further leverages language models to enhance diffusion-based models for open-vocabulary object detection. Diffusion models are also considered for 3D object detection~\cite{XU2024_3DiffTection} in the context of LiDAR-Camera fusion \cite{Xiang20233_DifFusionDetDM} and other tasks such as pose estimation \cite{Tan2024_DiffRegPose} and semantic segmentation \cite{Gu2024_DiffusionInst,Amit2022_segdiffimagesegmentationdiffusion}.

\section{Preliminary}
\label{sec:preliminary}

\begin{figure}
    \centering
    \includegraphics[width=0.9\linewidth]{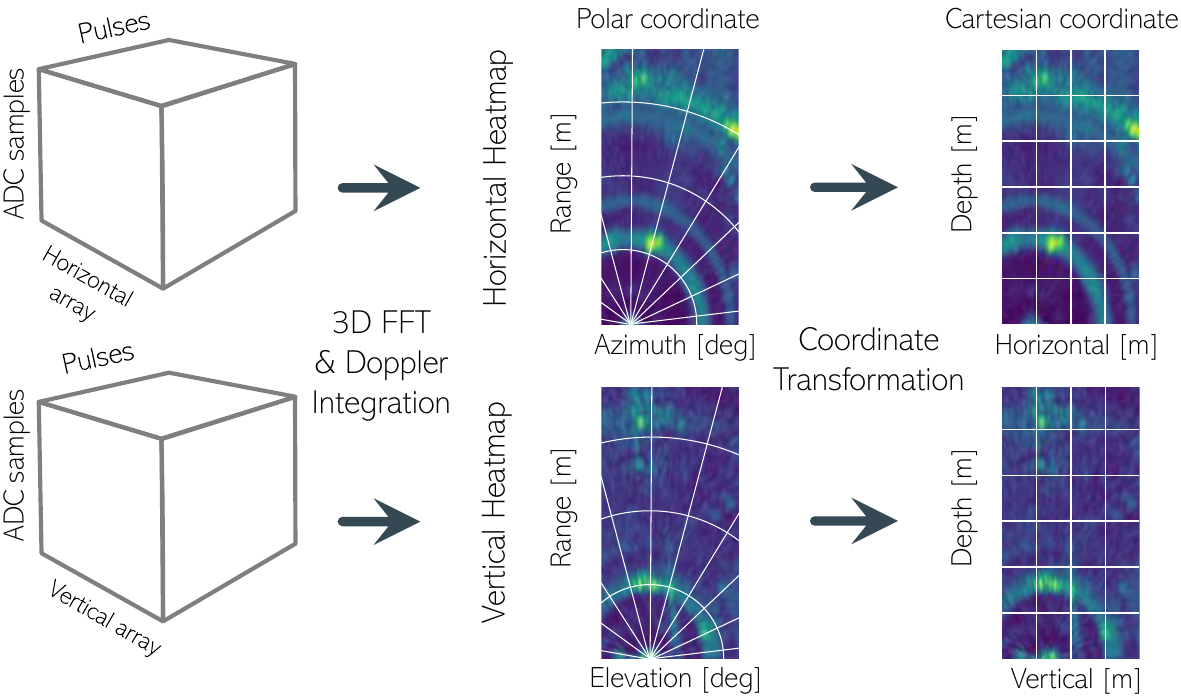}
    \vspace{-3mm}
    \caption{Generation of multi-view heatmaps from raw data.}
    \label{fig:radar}
    \vspace{-6mm}
\end{figure}

\paragraph{Multi-View Radar Heatmaps}
\label{sec:preprocessing}
derive from raw data captured by horizontal and vertical radar arrays, where sampling reflected pulses across each array builds a 3D data cube of ADC samples, pulse samples and array elements (Fig.~\ref{fig:radar}). A 3D FFT transforms each cube into radar spectra along range, Doppler, and angle (azimuth or elevation). Integrating over Doppler yields 2D polar heatmaps (range–azimuth and range–elevation), which are then mapped to Cartesian coordinates. The resulting heatmaps for frame $m$ are denoted $\B{Y}_{\OP{hor}}(m) \in {\cal{R}}^{W\times D}$ and $\B{Y}_{\OP{ver}}(m) \in {\cal{R}}^{H\times D}$. Stacking $M$ consecutive frames gives $\B{Y}_{\OP{hor}}\in {\cal{R}}^{M\times W\times D}$ and $\B{Y}_{\OP{ver}}\in {\cal{R}}^{M\times H\times D}$ for temporal modeling.

\paragraph{Diffusion Models} such as the denoising diffusion probabilistic model (DDPM)~\cite{Ho2020_DDPM} and the denoising diffusion implicit model (DDIM)~\cite{Song2021_DDIM}, define Markovian or non-Markovian forward processes by gradually adding noise to samples $\B{x}_0$, e.g., image pixels,
\begin{equation}
    \label{eq:direct_sampling}
    q\left(\B{x}_t \mid \B{x}_0\right)=\C{N}\left(\B{x}_t ; \sqrt{\bar{\alpha}_t} \B{x}_0,\left(1-\bar{\alpha}_t\right) \B{I}\right),
\end{equation}
where $t \in\{0, \ldots, T\}$ and $\bar{\alpha}_t=\prod_{s=0}^t\left(1-\beta_s\right)$ with $\beta_s$ denoting the noise variance schedule. At time $t$, $\B{x}_t=\sqrt{\bar{\alpha}_t} \B{x}_0 + \sqrt{\left(1-\bar{\alpha}_t\right) } \B{\epsilon}$ with $\B{\epsilon} \sim \C{N}\left(\B{0},  \B{I}\right)$. 

During training, a noise prediction network is trained to estimate $\B{\epsilon}$ from $\B{x}_t$ and time index $t$ by minimizing $\min_{\theta} \| \B{\epsilon}_{\theta} (\B{x}_t, t) - \B{\epsilon} \|^2$, where $\theta$ represents the trainable weights. During inference, a random $\B{x}_T$ is drawn from the standard Gaussian distribution and iteratively denoised using the trained $\B{\epsilon}_{\theta} (\B{x}_t, t)$ in reverse time:  $\B{x}_T \rightarrow \cdots \rightarrow \B{x}_{t} \rightarrow \B{x}_{t-1} \rightarrow \cdots \rightarrow \B{x}_0$. Sampling strategies such as DDPM \cite{Ho2020_DDPM} 
\begin{equation}
\label{eq:sampling}
    \B{x}_{t-1} =  \left( \B{x}_{t} - \frac{1-\alpha_t}{\sqrt{1-\bar{\alpha}_t}} \B{\epsilon}_{\theta} (\B{x}_t, t) \right)/\sqrt{\alpha_t}  + \sigma_t \B{\epsilon}_t,
\end{equation}
with $\B{\epsilon}_t \sim \C{N}\left(\B{0},  \B{I}\right)$, and DDIM \cite{Song2021_DDIM} can be used to trade off between quality and speed. More details are in Appendix~\ref{sec:detailedDiffusion}. 

\section{REXO: BBox Diffusion in 3D Radar Space}
\label{sec:proposed}

\begin{figure}[t]
    \centering
    \includegraphics[width=\linewidth]{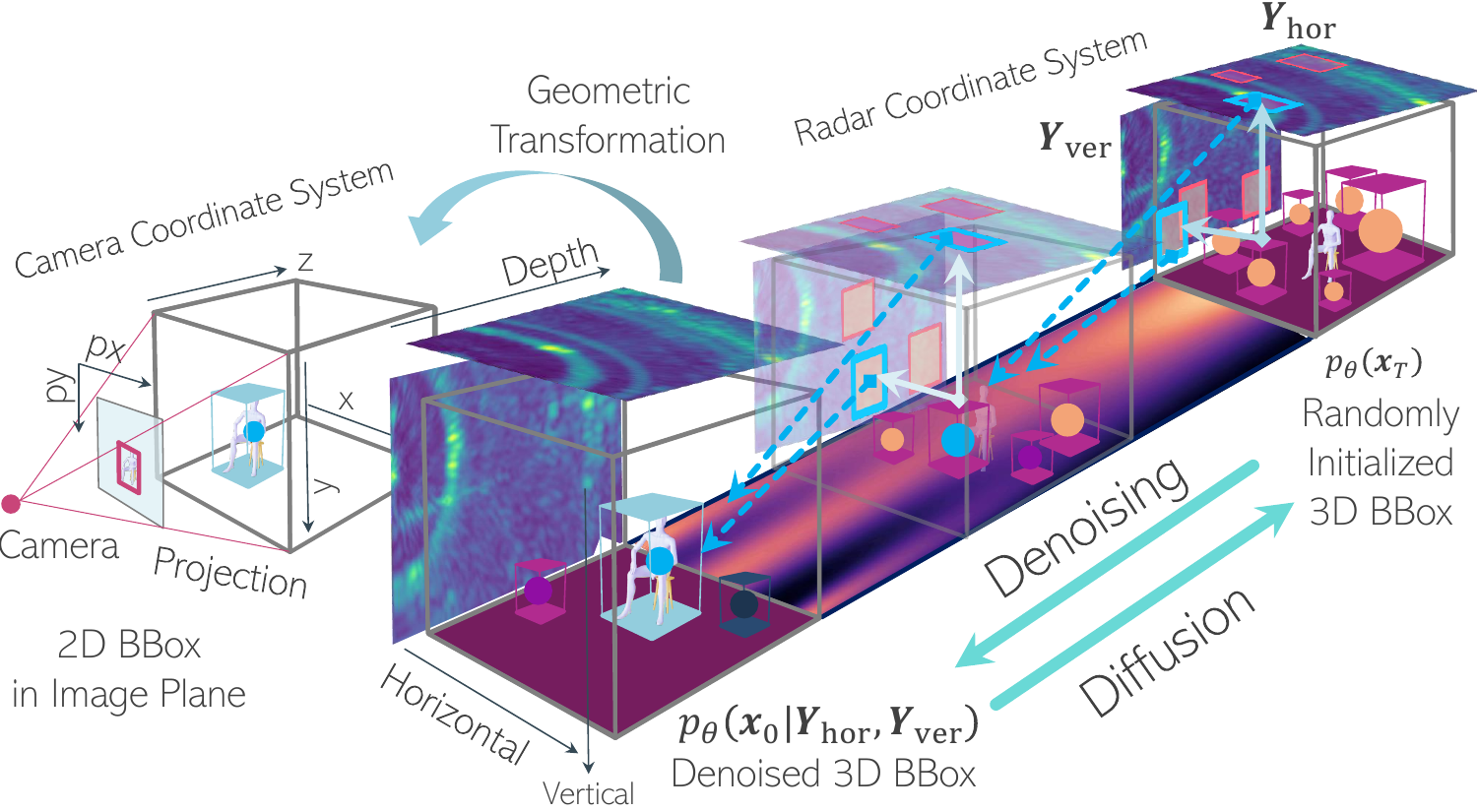}
    \vspace{-7mm}
    \caption{REXO: 1) 3D BBox diffusion process in the radar space; 2) Geometric transformation and 3D-to-2D projection onto the image plane for geometry-aware supervision. 
    }
    \label{fig:concept}
    \vspace{-6mm}
\end{figure}

DiffusionDet~\cite{Chen2023_diffusiondet} reformulates object detection as a denoising diffusion process, treating $\B{x}_t$ as 2D BBox parameters instead of image pixels. As shown in Fig.~\ref{fig:concept}, we extend this to multi-view radar by lifting $\B{x}_t$ to 3D BBoxes in radar coordinates: $\B{x}_{t} = \LM{c^t_{x}, c^t_{y}, c^t_{z}, w^t, h^t, d^t}^{\top}\in \R^6$, where $\LS{c^t_{x}, c^t_{y}, c^t_{z}}$ defines the center and $\LS{w^t, h^t, d^t}$ the size at time $t$ in the Cartesian $\{\OP{horizontal, vertical, depth}\}$ space. Conditioned on radar heatmaps $\{\B{Y}_{\OP{hor}},\B{Y}_{\OP{ver}}\}$, REXO performs 3D BBox diffusion in two phases (Fig.~\ref{fig:concept}): 
1) a \textbf{forward process} that adds noise to ground-truth (GT) BBoxes $\B{x}_0$ to produce random $\B{x}_T$ during training, and 
2) a \textbf{reverse process} that denoises random $\B{x}_T$ to estimate noise-free $\hat{\B{x}}_0$ during inference. The denoised BBoxes are also projected to the 2D image plane for supervision in both radar and image domains. We describe REXO in four parts: training, inference, ground-level constraint and loss.

\subsection{Training}
\label{sec:training}
\begin{figure*}[t]
    \centering
    \includegraphics[width=\linewidth]{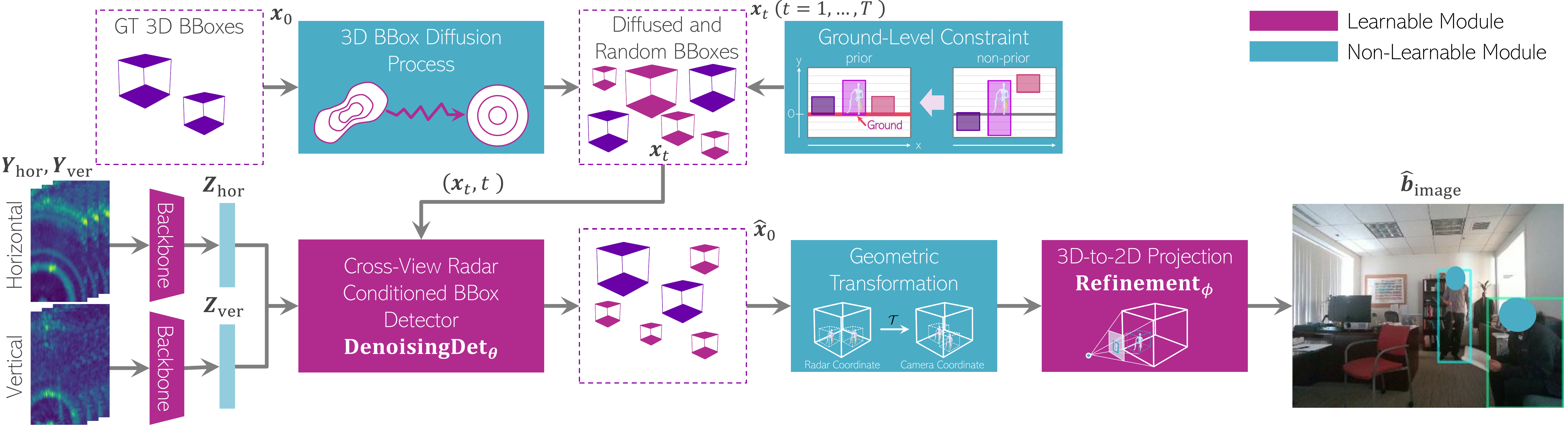}
    \vspace{-6mm}
    \caption{
    \textbf{REXO training:}
    1) A shared backbone extracts horizontal/vertical radar features $\{\B{Z}_{\OP{hor}},\B{Z}_{\OP{ver}}\}$;
    2) Ground‑truth 3D BBoxes $\B{x}_0$ are diffused to noisy $\B{x}_t$;
    3) $\B{x}_t$ is grounded using a ground-level constraint;
    4) $\OP{DenoisingDet}_{\theta}$ projects $\B{x}_t$ onto both views and uses the aligned features to recover $\hat{\B{x}}_0$;
    5) A radar‑to‑camera transform and 3D-to-2D projection yield image BBoxes $\hat{\B{b}}_{\OP{image}}$, enabling geometry‑aware supervision in radar space and image plane.
    }
    \label{fig:architecture_train}
    \vspace{-6mm}
\end{figure*}

We describe REXO training, as illustrated in Fig.~\ref{fig:architecture_train}. 

\paragraph{Backbone:} Taking the two radar heatmaps $\B{Y}_{\OP{hor}}$ and  $\B{Y}_{\OP{ver}}$ as inputs, a backbone network (e.g., ResNet~\cite{He2016_ResNet}) generates horizontal-view and vertical-view radar feature maps separately: $\B{Z}_{\OP{hor}} = \OP{backbone}\LS{\B{Y}_{\OP{hor}}}$ and $\B{Z}_{\OP{ver}} = \OP{backbone}\LS{\B{Y}_{\OP{ver}}}$, where learnable parameters in the $\OP{backbone}$ are shared across both views. Each feature map is generated as $L$ multi-scale feature maps in $\R^{C\times \frac{W}{sl}\times \frac{D}{sl}}$ or $\R^{C\times \frac{H}{sl}\times \frac{D}{sl}}$ by using feature pyramid network~\cite{Lin2017_fpn} where $C$, $s$ and $l\in\LM{1,\cdots,L}$ represent the number of channels, downsampling ratio over the spatial dimension and the pyramid level, respectively.

\paragraph{Initialization of $\B{x}_0$ and Forward Process to $\B{x}_t$:}  For a given number of BBoxes $N_{\OP{train}}$ to be detected, $\B{x}_{0}$ is simply initialized by the 3D BBox GT in the radar space $\B{x}_{\OP{radar}} = \LM{c_{x}, c_{y}, c_{z}, w, h, d}^{\top}\in \R^6$ and padded with random 3D BBoxes $\B{x}_{\OP{rand}}\sim\C{N}\LS{\B{0}, \B{I}_6}$ if $N_{\OP{train}} > N_{\OP{GT}}$. The diffused 3D BBox $\B{x}_{t}$ at time $t$ can be generated as 
\begin{equation} \label{xt}
    \B{x}_t = \sqrt{\bar{\alpha}_t} \B{x}_0 + \sqrt{1-\bar{\alpha}_t} \B{\epsilon},
\end{equation}
where $\B{\epsilon} \sim \C{N}\LS{\B{0}, \B{I}_6}$ and $\bar{\alpha}_t$ is defined in Section~\ref{sec:preliminary}. 

\paragraph{Cross-View Radar-Conditioned BBox Detector $\OP{DenoisingDet}_{\theta}$} includes explicit cross-view feature association and radar-conditioned 3D BBox detector. 

1) Explicit cross-view feature association: Given the noisy 3D BBox $\B{x}_{t}$ in \eqref{xt}, the $\B{x}_t$-guided cross-view feature association first projects $\B{x}_{t}$ onto the two radar views, resulting in two 2D BBoxes (purple solid lines of Fig.~\ref{fig:comparison_obj_detection} (d)), 
\begin{align} \label{eq:BBox_projection}
    \B{x}_{t, \OP{hor}} = \{c^t_{x}, c^t_{z}, w^t, d^t\}^{\top},
    \B{x}_{t, \OP{ver}} = \{c^t_{y}, c^t_{z}, h^t, d^t\}^{\top},
\end{align}
and then crops out the cross-view 2D radar features
\begin{align} \label{eq:roialign}
    \B{Z}_{\OP{hor}}^{\OP{crop}} & = {\OP{RoIAlign}}(\B{Z}_{\OP{hor}}, \B{x}_{t, \OP{hor}})
    \in\R^{C \times r \times r}, \notag  \\
    \B{Z}_{\OP{ver}}^{\OP{crop}} & = {\OP{RoIAlign}}(\B{Z}_{\OP{ver}}, \B{x}_{t, \OP{ver}})
    \in\R^{ C \times r \times r},
\end{align}
via a standard ROIAlign operation~\cite{He2017_maskrcnn},  where $r$ denotes a fixed spatial resolution, e.g., $r=7$. At time $t$, this process yields $N_{\OP{train}}$ pairs of associated radar features 
\begin{equation}
\label{eq:feature_association}
    \B{Z}_{\OP{radar}}^{\OP{crop}} = \LM{\B{Z}_{\OP{hor}}^{\OP{crop}}, \B{Z}_{\OP{ver}}^{\OP{crop}}} \in\R^{ C \times r \times 2r},
\end{equation} 
each corresponding to a noisy 3D BBox $\B{x}_t$. 

2) Radar-conditioned 3D BBox detector: Conditioned on $\B{Z}_{\OP{radar}}^{\OP{crop}}$, a BBox detector with learnable weights $\theta$ is trained to estimate the BBox $\hat{\B{x}}_0$ and the class scores $\hat{\B{p}}$ as
\begin{equation} \label{b0hat}
    \LM{\hat{\B{x}}_0, \hat{\B{p}}} = \OP{BBoxDet}_{\theta}\left(\B{Z}_{\OP{radar}}^{\OP{crop}}, t\right),
\end{equation}
where $t$ specifies the timestep embedding. In our indoor setting, we use a two-class softmax over \{\emph{person, background}\}. The class-head can extend to $C$ classes (including background) by using a $C$-way softmax with cross-entropy.

Grouping all the steps from \eqref{eq:BBox_projection} to \eqref{b0hat} results in the $\OP{DenoisingDet}_{\theta}$ module of Fig.~\ref{fig:architecture_train}:
\begin{equation}
    \label{eq:detector}
    \LM{\hat{\B{x}}_0, \hat{\B{p}}} = \OP{DenoisingDet}_{\theta}\left( \B{x}_{t}, t, \B{Z}_{\OP{hor}},\B{Z}_{\OP{ver}}\right),
\end{equation}
where all trainable weights $\theta$ are inherited from the BBox detector $\OP{BBoxDet}_{\theta}$. Further details are in Appendix~\ref{sec:denoisingDet}.

\paragraph{3D-to-2D Projection with Learnable Refinement.}
REXO further projects $\hat{\B{x}}_0$ in \eqref{eq:detector} into the 2D image plane through the 3D camera coordinate system via a calibrated geometric transformation $\C{T}$. By setting $\hat{\B{x}}_{\OP{radar}} = \hat{\B{x}}_{0}$, we convert each of the $8$ corners of the corresponding 3D BBox $\hat{\B{x}}_{\OP{radar}}$ using
\begin{equation}
    \label{c2r}
    \B{x}_{\OP{camera}}^{i} = \B{R}\hat{\B{x}}_{\OP{radar}}^{i} + \B{v}, \quad i=1,2, \cdots, 8,
\end{equation}
where $\hat{\B{x}}_{\OP{radar}}^{i}$ is the $i$-th corner of $\hat{\B{x}}_{\OP{radar}}$, $\B{R}$ is the calibrated 3D rotation matrix, and $\B{v}$ is the calibrated translation vector. Each 3D corner $\B{x}_{\OP{camera}}^{i}$ is projected to the image plane through the calibrated pinhole model, and the extrema of the eight projected points yield the initial box $\B{b}_{\OP{init}}$
\begin{align}
    \label{eq:image_plane_projection}
    \B{b}_{\OP{init}} = \LM{\bar{c}_{x}, \bar{c}_{y}, \bar{w}, \bar{h}}^{\top} =\OP{proj}_{\OP{init}}\LS{\B{x}_{\OP{camera}}}.
\end{align}
Since $\B{b}_{\OP{init}}$ systematically overshoots the ground‑truth extent (see Appendix~\ref{sec:details_refinement_module}), we attach a refinement module with learnable parameter $\B{\phi}$ to obtain the offset:
\begin{equation}
    \Delta\B{b} = \LM{\Delta{\bar{x}}, \Delta{\bar{y}}, \Delta{\bar{w}}, \Delta{\bar{h}}}^{\top} = \OP{Refinement}_{\B{\phi}}\LS{\B{f}},
\end{equation}
where $\B{f}=\OP{Predictor}\LS{\OP{e}_t,\B{Z}_{\OP{radar}}^{\OP{crop}}}$ is the time-dependent feature. $\OP{e}_t$ denotes the timestep embedding~\cite{Ho2020_DDPM} and $\OP{Predictor}$ denotes the time-dependent predictor~\cite{Chen2023_diffusiondet} with the radar feature and the embedding (see Appendix~\ref{sec:denoisingDet} for more details). Applying these offsets produces the final image‑plane box $\widehat{\B{b}}_{\OP{image}}$, achieving tighter alignment without sacrificing geometric consistency.
\begin{equation}
\label{eq:supervision}
   \widehat{\B{b}}_{\OP{image}} = \{ \bar{c}_{x}+\bar{w}\Delta{\bar{x}}, \bar{c}_{y}+\bar{h}\Delta{\bar{y}}, e^{\Delta{\bar{w}}}\bar{w}, e^{\Delta{\bar{h}}}\bar{h} \}^{\top}.
\end{equation}

\begin{table*}[t]
    \footnotesize
    \centering
    \caption{Evaluation on $4$ data splits of the MMVR. The gray hatch represents the best performance for each metric.}
    \vspace{-3mm}
    \setlength\tabcolsep{7.0pt}
    \begin{tabular}{l rrr r rrr r rrr r rrr}
        \toprule
        \multirow{2}{*}{$\OP{Method}$} & \multicolumn{3}{c}{P1S1} && \multicolumn{3}{c}{P1S2} && \multicolumn{3}{c}{P2S1} && \multicolumn{3}{c}{P2S2}\\
        \cline{2-4} \cline{6-8} \cline{10-12} \cline{14-16}
         & \multicolumn{1}{c}{$\OP{AP}$} & \multicolumn{1}{c}{$\OP{AP_{50}}$} & \multicolumn{1}{c}{$\OP{AP_{75}}$} && \multicolumn{1}{c}{$\OP{AP}$} & \multicolumn{1}{c}{$\OP{AP_{50}}$} & \multicolumn{1}{c}{$\OP{AP_{75}}$} && \multicolumn{1}{c}{$\OP{AP}$} & \multicolumn{1}{c}{$\OP{AP_{50}}$} & \multicolumn{1}{c}{$\OP{AP_{75}}$} && \multicolumn{1}{c}{$\OP{AP}$} & \multicolumn{1}{c}{$\OP{AP_{50}}$} & \multicolumn{1}{c}{$\OP{AP_{75}}$} \\
        \midrule
            RFMask & 25.53 & 67.30 & 15.86 &  & 24.46 & 66.82 & 11.22 &  & 31.37 & 61.50 & 27.48 &  & 6.03 & 22.77 & 0.88 \\
            RFMask3D & 34.84 & 69.57 & 31.74 & & 30.75 & 76.48 & 16.23 & & 39.89 & 80.38 & 35.35 & & 12.26 & 37.01 & 4.34 \\
            DETR & 35.64 & 77.59 & 28.00 &  & 28.51 & 75.90 & 13.42 &  & 29.53 & 63.08 & 25.35 &  & 9.29 & 34.69 & 2.49 \\
            RETR & \cellcolor{gray!20}39.62 & \cellcolor{gray!20}80.55 & 33.84 &  & 30.16 & 78.95 & 15.17 &  & 46.75 & 83.80 & 46.06 &  & 12.45 & 41.30 & 4.96 \\
            \midrule
            \textbf{REXO} & 39.23 & 73.46 & \cellcolor{gray!20}37.83 &  & \cellcolor{gray!20}36.48 & \cellcolor{gray!20}87.02 & \cellcolor{gray!20}20.51 &  & \cellcolor{gray!20}48.35 & \cellcolor{gray!20}85.89 & \cellcolor{gray!20}48.38 &  & \cellcolor{gray!20}23.47 & \cellcolor{gray!20}64.41 & \cellcolor{gray!20}10.44 \\
        \bottomrule
    \end{tabular}
    \label{tab:main_mmvr}
    \vspace{-3mm}
\end{table*}

\subsection{Inference}
\label{sec:inference}
REXO infers objects by reversing the diffusion process. Given a target count $N$, we sample random 3D boxes $\B{x}_T \sim\C{N}\LS{\B{0}, \B{I}_6}$ in the radar coordinate system at $t=T$ and denoise them down to $t=1$.

\paragraph{Denoising Process in 3D Radar Space:}
With $\B{x}_{t}$ and radar features $\{\B{Z}_{\OP{hor}},\B{Z}_{\OP{ver}}\}$, the trained $\OP{DenoisingDet}_{\theta}$ in~\eqref{eq:detector} predicts $\hat{\B{x}}_{0}$, giving
\begin{align} 
    p_\theta &\left(\B{x}_{t-1} \mid \B{x}_t, \B{Z}_{\OP{hor}},\B{Z}_{\OP{ver}}\right) = \C{N}(\sqrt{\alpha_{t-1}} \B{x}_0 + \gamma \B{\epsilon}_{\theta}^{\LS{t}}, \sigma_t^2 \B{I}_6 ), \nonumber
\end{align}
\begin{align}
    \label{eq:reverse}
    \B{x}_{t-1} = \sqrt{\alpha_{t-1}} \hat{\B{x}}_{0} + \sqrt{1-\alpha_{t-1}-\sigma_{t}^{2}} \cdot \B{\epsilon}_{\theta}^{(t)} + \sigma_t \B{\epsilon}_t,
\end{align}
where $\B{\epsilon}_{\theta}^{\LS{t}} = {\LS{\B{x}_t-\sqrt{\alpha_t} \hat{\B{x}}_0}}/{\sqrt{1-\alpha_t}}$ specifies the direction pointing to the noisy BBox $\B{x}_t$ at time $t$, and $\B{\epsilon}_t \sim \C{N}(\B{0}, \B{I}_6)$ represents a random BBox. Note that the denoising step is inherently conditioned on the cross-view radar feature maps via the estimated $\hat{\B{x}}_0$ from the $\OP{DenoisingDet}_{\theta}$ module.

\paragraph{2D Image Plane BBox Prediction:} 
After the final step, $\B{x}_{0}$ ($=\hat{\B{x}}_{\OP{radar}}$) is converted to image plane boxes $\hat{\B{b}}_{\OP{image}}$ via the radar–to-camera transform in~\eqref{c2r} and the 3D-to-2D projection of~\eqref{eq:supervision}. Boxes whose class scores exceed a threshold are output as detections.

\subsection{Ground-Level Constraint}
\label{sec:prior_constrained_grounding}
Since the BBoxes are now explicitly defined in the 3D radar coordinate system, it is natural to incorporate prior knowledge as a constraint into the diffusion process. Unlike DiffusionDet and RETR, we enforce the reduced five 3D parameters by grounding with $h^t/2$, allowing 3D and 2D gradients to flow jointly and guiding the denoising process under strict geometric constraints. This ensures that objects are correctly positioned on the floor, reflecting realistic spatial relationships (see the Ground-Level Constraint in Fig.~\ref{fig:architecture_train}):
\begin{equation}
    \label{eq:grounded_bbox}
    \B{x}_t = \LM{c^t_{x}, h^t/2, c^t_{z}, w^t, h^t, d^t}^{\top}.
\end{equation}
Using this constrained $\B{x}_t$ as in \eqref{xt}, REXO predicts $N_{\OP{train}}$ 3D BBoxes $\hat{\B{x}}_{\OP{radar}}$ and 2D BBoxes $\hat{\B{b}}_{\OP{image}}$, while preserving geometric consistency.

\subsection{Loss Function}
\label{sec:hybrid_loss}
To ensure consistency between the radar and image plane representations, we adopt a simplified scheme of the Tri-plane loss~\cite{Yataka2024_retr} that directly calculates the loss of 3D BBox. REXO employs the Hungarian match cost~\cite{kuhn1955_hungarian} with a geometry-aware loss function $\C{L}_{\OP{box}}^{\OP{GA}}$ computed in both the 3D and 2D spaces:
\begin{equation}
    \label{eq:loss}
    \C{L}_{\OP{box}}^{\OP{GA}} 
    =  \;\lambda_{\OP{3D}}\C{L}_{\OP{box}}^{\OP{3D}}\LS{\B{x}_{\OP{radar}}, \widehat{\B{x}}_{\OP{radar}}} 
    + \lambda_{\OP{2D}}\C{L}_{\OP{box}}^{\OP{2D}}(\B{b}_{\OP{image}}, \widehat{\B{b}}_{\OP{image}}),
\end{equation}
where the 3D/2D BBox loss is defined as $\C{L}_{\OP{box}}^{*}(\B{x}, \widehat{\B{x}}) = \lambda_{\OP{GIoU}} \C{L}_{\OP{GIoU}}(\B{x}, \widehat{\B{x}}) + \lambda_{\OP{L_1}} \C{L}_{\OP{L_1}}(\B{x}, \widehat{\B{x}})$ representing a weighted combination of the generalized intersection over union (GIoU) loss $\C{L}_{\OP{GIoU}}$~\cite{Rezatofighi2019_giou} and the $\ell_1$ loss $\C{L}_{\OP{L_1}}$, and the coefficients $\lambda$ balance the relative contribution of each loss term. REXO determines the optimal assignment $\sigma^*_{\OP{GA}}$ by minimizing the matching cost that combines the original classification cost $\C{L}_{\OP{class}}$ and $\C{L}_{\OP{box}}^{\OP{GA}}$. 

\section{Experiments}
\label{sec:experiments}
We demonstrate the effectiveness of REXO through evaluations on two open radar datasets: HIBER ~\cite{Wu2023_RFMask} and MMVR~\cite{Perry2024_MMVR}. 

\subsection{Setup}
\paragraph{High-Resolution Indoor Radar Datasets:} MMVR includes multi-view radar heatmaps collected from over $25$ human subjects across $6$ rooms over a span of $9$ days. It consists of 345K data frames collected in 2 protocols: 1) Protocol 1 (P1: Open Foreground) with $107.9$K frames in an open-foreground room with a single subject; and 2) Protocol 2 (P2: Cluttered Space) with $237.9$K frames in $5$ cluttered rooms with single and multiple subjects. Under each protocol, two data splits are defined to evaluate radar perception performance: 1) S1: a random data split and 2) S2: a cross-session, unseen split.

HIBER, partially released, includes multi-view radar heatmaps from $10$ human subjects in a single room but from different angles with multiple data splits. In our evaluation,  we used the ``WALK'' data split, consisting of $73.5$K data frames with one subject walking in the room. For both datasets, annotations such as 3D BBoxes in the radar coordinate system and 2D image plane BBoxes are provided to train the baseline methods and REXO. 

\paragraph{Implementation:}
We consider RFMask~\cite{Wu2023_RFMask}, DETR~\cite{Carion2020_detr} and RETR~\cite{Yataka2024_retr} as baseline methods. Additionally, we evaluate a 3D extension of RFMask, referred to as RFMask3D (see Appendix~\ref{sec:rfmask3D} for details), that takes the two radar views as inputs for BBox prediction. 
Since RFMask and DETR originally compute the BBox loss only in the 2D horizontal  radar plane and the 2D image plane, respectively, we follow the implementation of RETR and enhance both methods with a unified bi-plane BBox loss. Furthermore, we introduce a DETR variant with a top-$K$ feature selection, allowing it to take features from both horizontal and vertical heatmaps as input. 
For RETR, we set the number of object queries to $10$. To ensure a fair comparison, we also set $N_{\OP{train}}=10$ for REXO during training. 
All methods are evaluated using $M=4$ consecutive radar frames.
Additional hyperparameter settings are provided in Appendix~\ref{sec:experimental_setting}.

\paragraph{Metrics:} We evaluate performance using average precision (AP) at two IoU thresholds of $0.5$ ($\OP{AP_{50}}$) and $0.75$ ($\OP{AP_{75}}$), along with the mean AP ($\OP{AP}$) computed over thresholds in the range of $[0.5: 0.05:0.95]$. For detailed metric definitions, refer to Appendix~\ref{sec:definition_metrics}. 

\subsection{Main Results}

\begin{table}[t]
    \footnotesize
    \centering
    \caption{Evaluation on the WALK data split of the HIBER.}
    \vspace{-3mm}
    \setlength\tabcolsep{17.2pt}
    \begin{tabular}{l rrr}
        \toprule
        \multirow{2}{*}{$\OP{Method}$} & \multicolumn{3}{c}{WALK} \\
        \cline{2-4}
         & \multicolumn{1}{c}{$\OP{AP}$} & \multicolumn{1}{c}{$\OP{AP_{50}}$} & \multicolumn{1}{c}{$\OP{AP_{75}}$}  \\
        \midrule
        RFMask & 17.77 & 52.46 & 6.78  \\
        RFMask3D & 16.58 & 48.10 & 6.53  \\
        DETR & 14.45 & 47.33 & 4.25  \\
        RETR & 22.09 & 59.83 & 10.99  \\
        \midrule
        \textbf{REXO} & \cellcolor{gray!20}25.33 & \cellcolor{gray!20}62.55 & \cellcolor{gray!20}15.83  \\
        \bottomrule
    \end{tabular}
    \label{tab:main_hiber}
    \vspace{-3mm}
\end{table}
\begin{figure}[t]
    \centering
    \begin{subfigure}[b]{0.49\linewidth}
        \centering
        \includegraphics[width=0.9\linewidth]{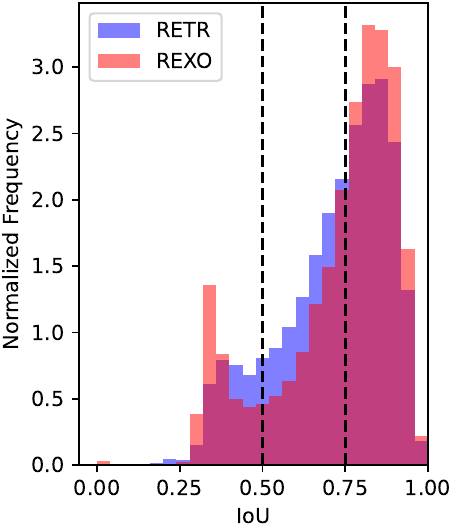}
        \vspace{-2mm}
        \caption{P1S1 }
        \label{fig:iou_histograms_p1s1}
    \end{subfigure}
    \hfill
    \begin{subfigure}[b]{0.49\linewidth}
        \centering
        \includegraphics[width=0.9\linewidth]{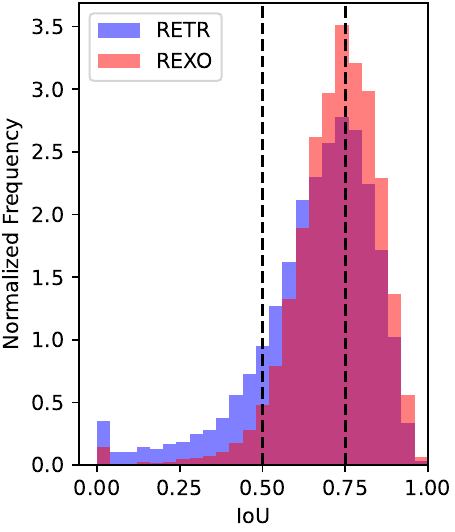}
        \vspace{-2mm}
        \caption{P1S2}
        \label{fig:iou_histograms_p1s2}
    \end{subfigure}
    \hfill
    \vspace{-3mm}
    \caption{AP breakdowns with IoU histograms on MMVR.}
    \label{fig:iou_histograms_p1}
    \vspace{-6mm}
\end{figure}

\begin{figure}[t]
    \begin{minipage}[b]{0.27\textwidth}
        \footnotesize
        \setlength\tabcolsep{2pt}
        \begin{tabular}{c|cc}
            \toprule
            \multirow{1}{*}{$\OP{Grounding}$} & \multicolumn{1}{c}{$\OP{AP}$ (P2S2)} & \multicolumn{1}{c}{$\OP{AP}$ (WALK)}  \\
            \midrule
            $\times$ & 22.67 & 21.11 \\
            \checkmark    & \cellcolor{gray!20}23.47 & \cellcolor{gray!20}25.33 \\
            \bottomrule
        \end{tabular}
        \vspace{-0.13in}
        \captionof{table}{
            The ground-level constraint can improve the detection performance on both datasets.
        }
        \label{tab:grounding}
    \end{minipage}
    \hfill
    \begin{minipage}[b]{0.18\textwidth}
        \footnotesize
        \setlength\tabcolsep{11pt}
        \begin{tabular}{c|cc}
            \toprule
            $D$ [cm] & $\OP{AP}$ \\
            \midrule
            $D \leq 20$ & 9.93\\
            $D > 20$ & \cellcolor{gray!20}23.47 \\
            \bottomrule
        \end{tabular}
        \vspace{-0.13in}
        \captionof{table}{AP drops when depth difference $D$ is less than 20 cm.}
        \label{tab:ap_drop_same_depth}
    \end{minipage}
    \vspace{-4mm}
\end{figure}

\begin{table*}[t]
    \footnotesize
    \subfloat[
        Strong \textbf{2D ($\lambda_{\OP{2D}}$)/3D ($\lambda_{\OP{3D}}$) supervision} improves performance.
    \label{tab:loss_weight}
    ]{
    \setlength\tabcolsep{4pt}
    \begin{minipage}{0.15\linewidth}{
        \centering
        \begin{tabular}{ccc}
        \toprule
        
        $\lambda_{\OP{3D}}$ & $\lambda_{\OP{2D}}$ & \multicolumn{1}{c}{$\OP{AP}$} \\
    
        \midrule
        0.00 & 1.00 & 0.98  \\
        0.50 & 1.00 & 4.23  \\
        1.00 & 0.10 & 15.55 \\
        1.00 & 0.50 & 19.38 \\
        1.00 & 1.00 & \cellcolor{gray!20}23.47  \\
        \\ 
        
        \bottomrule
        \end{tabular}}
    \end{minipage}
    }
    \hspace{1em}
    \subfloat[
        \# \textbf{of BBoxes for training}. REXO remains stable.
    \label{tab:num_bboxes_training}
    ]{
    \setlength\tabcolsep{1pt}
    \begin{minipage}{0.15\linewidth}{\begin{center}
        \centering
        \begin{tabular}{ccc}
        \toprule        
        \multicolumn{1}{c}{$\OP{Method}$} & $N_{\OP{train}}$ & \multicolumn{1}{c}{$\OP{AP}$} \\
        \midrule
        RETR & 10 & \cellcolor{gray!20}12.45 \\
         & 20 & 9.85 \\
         & 50 & 8.49 \\
        REXO & 10 & \cellcolor{gray!20}23.47 \\
         & 20 & 20.94 \\
         & 50 & 19.67 \\
        \bottomrule
        \end{tabular}
        \end{center}}
    \end{minipage}
    }
    \vspace{-1mm}
    \hspace{1em}
    \subfloat[
        \# \textbf{of BBoxes for inference}. REXO sustains its $\OP{AP}$ as $N$ increases. 
    \label{fig:comp_n_bbox_inference}
    ]{
    \setlength\tabcolsep{4.5pt}
    \begin{minipage}{0.16\linewidth}{\begin{center}
        \centering
        \begin{tabular}{ccc}
        \toprule
        $N$ & REXO & RETR \\
        \midrule
        2 & 8.87 & 8.36 \\
        10 & \cellcolor{gray!20}23.48 & \cellcolor{gray!20}12.45 \\
        20 & 23.00 & 6.57 \\
        40 & 22.32 & 3.63 \\
        60 & 21.94 & 2.65 \\
        80 & 21.70 & 2.16 \\
        \bottomrule
        \end{tabular}
        \end{center}}
    \end{minipage}
    }
    \vspace{-1mm}
    \hspace{1em}
    \subfloat[
        \# \textbf{of denoising steps}. More steps slightly improve detection.
    \label{fig:comp_diff_steps}
    ]{
    \setlength\tabcolsep{9pt}
    \begin{minipage}{0.16\linewidth}{\begin{center}
        \begin{tabular}{cc}
        \toprule
        $\OP{Steps}$ & \multicolumn{1}{c}{$\OP{AP}$} \\
         \midrule
        1 & 23.48 \\
        3 & 24.01 \\
        5 & 24.12 \\
        7 & 24.17 \\
        9 & 24.25 \\
        10 & \cellcolor{gray!20}24.27 \\
        \bottomrule
        \end{tabular}
        \end{center}}
    \end{minipage}
    }
    \hspace{1em}
    \subfloat[
        \textbf{DiffusionDet vs. single-/multi-view REXO.} Multi-view achieves the best $\OP{AP}$.
    \label{tab:diffusiondet}
    ]{
    \setlength\tabcolsep{3pt}
    \begin{minipage}[b]{0.21\textwidth}{\begin{center}
        \begin{tabular}{lc}
            \toprule
            $\OP{Method}$ & $\OP{AP}$  \\
            \midrule
            DiffusionDet & 20.75  \\
            REXO (Horizontal) & 22.75  \\
            REXO (Vertical) & 7.18  \\
            REXO (Both Views)   & \cellcolor{gray!20}23.47  \\
            \\
            \\
            \bottomrule
        \end{tabular}
        \end{center}}
    \end{minipage}
    }
    \caption{Ablation study under P2S2 on MMVR.}
    \vspace{-5mm}
\end{table*}


\begin{figure}[t]
    \centering
    \begin{subfigure}[b]{0.49\linewidth}
        \centering
        \includegraphics[width=0.9\linewidth]{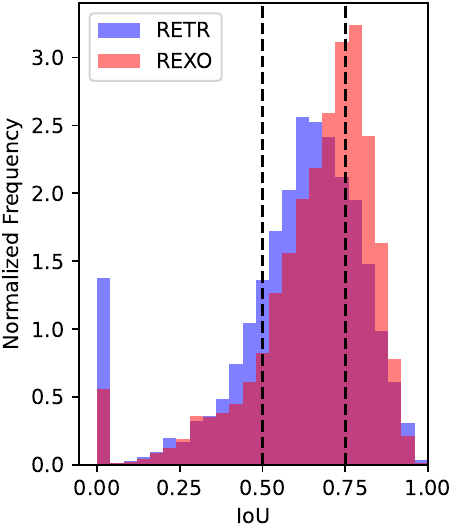}
        \vspace{-2mm}
        \caption{Seen Frames: REXO exhibits a similar histogram to that observed in P1S2.}
        \label{fig:iou_histograms_p2s2_seen}
    \end{subfigure}
    \hfill
    \begin{subfigure}[b]{0.48\linewidth}
        \centering
        \includegraphics[width=0.9\linewidth]{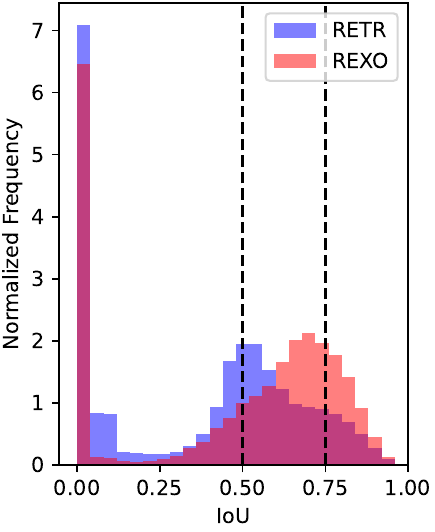}
        \vspace{-2mm}
        \caption{Unseen Frames: REXO has more high-quality predictions than RETR.}
        \label{fig:iou_histograms_p2s2_unseen}
    \end{subfigure}
    \vspace{-2mm}
    \caption{AP breakdowns with IoU histograms on MMVR.}
    \label{fig:iou_histograms_p2s2}
    \vspace{-6mm}
\end{figure}
\begin{figure*}[t]
    \centering
    \vspace{-2mm}
    \includegraphics[width=0.95\linewidth]{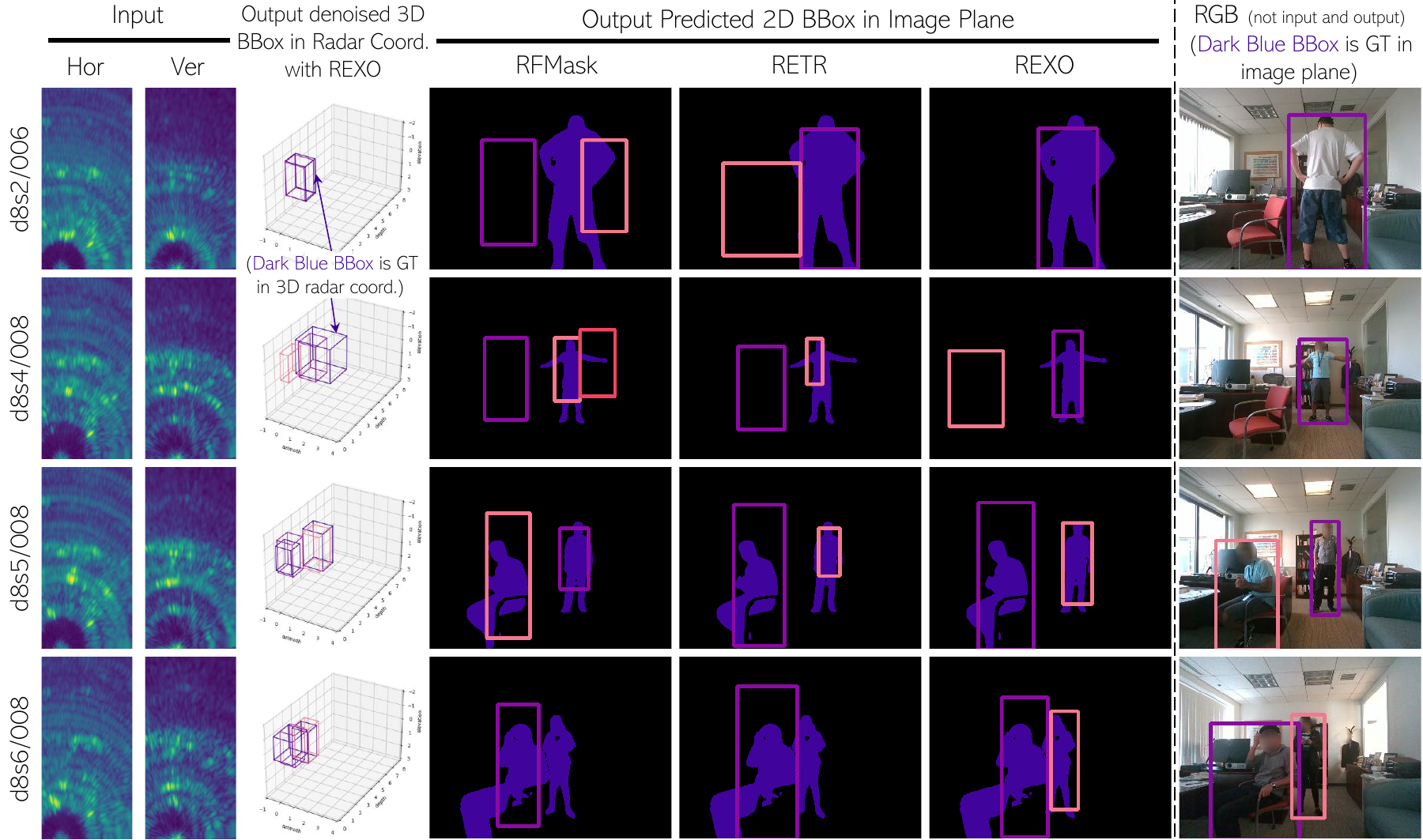}
    \caption{Visualization of unseen frames in P2S2 of MMVR: The left column shows the radar heatmaps, followed by the second column displaying predicted/GT 3D BBoxes in the radar space. Corresponding image-plane 2D BBox predictions are shown in the middle column for two baselines (RFMask and RETR) and REXO, with purple segmentation masks overlaid to illustrate the alignment with human GT. The right column presents the RGB images with GT 2D BBoxes for qualitative check.}
    \label{fig:visualization_main}
    \vspace{-4mm}
\end{figure*}

\begin{figure}[t]
    \centering
    \includegraphics[width=0.23\textwidth]{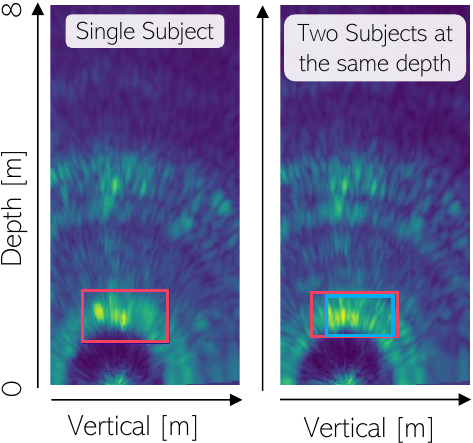}        
    \vspace{-3mm}
    \caption{Vertical radar heatmaps when a single subject (left) and two subjects (right) are at the same depth. }
    \label{fig:same_depth_heatmap}
    \vspace{-3mm}
\end{figure}

\paragraph{MMVR:}
Table~\ref{tab:main_mmvr} presents the results under the four combinations of  two protocols and two data splits of the MMVR dataset. REXO demonstrates significant performance improvements in P1S2, P2S1, and P2S2. Notably, in P2S2 where the test radar frames contain an entirely unseen environment during training, REXO outperforms the best baseline RETR by a large margin, boosting  $\OP{AP}$ from $12.45$ to $23.47$, highlighting its strong generalization capabilities. Surprisingly, under the simplest combination P1S1 where a single subject is recorded in the same room with a random data split, REXO’s performance is slightly lower than that of RETR, particularly on the metric $\OP{AP}_{50}$. 

To understand these differences, we break down the AP into IoU histograms for (a) P1S1 and (b) P1S2, as illustrated in Fig.~\ref{fig:iou_histograms_p1}, where blue and red histograms represent the IoU distributions for RETR and REXO, respectively, and the left and right dotted lines mark the two IoU thresholds at $0.5$ and $0.75$. It is seen that in Fig.~\ref{fig:iou_histograms_p1s1}, the excess of RETR over REXO (blue areas) over the IoU interval $[0.5, 0.75]$ is greater than that of REXO over RETR (pink areas) over the interval $[0.75, 1.0]$, explaining RETR's higher $\OP{AP}_{50}$ under P1S1. Meanwhile, REXO has better $\OP{AP}_{75}$ as it provides more high-quality predictions with IoU above $0.75$. 

\paragraph{HIBER:}
Table~\ref{tab:main_hiber} presents the results evaluated on the ``WALK'' data split of the HIBER dataset. For baselines, RFMask, RFMask3D, and DETR show comparable performance, while RETR exhibits the strongest baseline performance. REXO outperforms RETR across all evaluation metrics with an $\OP{AP}$ of $25.33$, surpassing RETR's $\OP{AP}$ at $22.09$. REXO attains $\OP{AP}_{50}$ of $62.55$ and $\OP{AP}_{75}$ of $15.83$, demonstrating strong performance in both low- and high-IoU BBox performance evaluations.  This ability  to consistently outperform the baselines across different IoU thresholds indicates REXO's robustness in capturing object localization with relatively better accuracy. 

\subsection{Ablation Study}
\label{sec:ablation}
We present ablation studies for REXO under the most challenging ``P2S2'' of the MMVR dataset.  Full results are provided in Appendix~\ref{sec:more_ablation}.

\paragraph{Effectiveness of Ground-Level Constraint:}
Table~\ref{tab:grounding} reports the effect of  ground-level constraint. In MMVR, the subject stands on the ground or sits in a chair, so the constraint is effective. The table shows that we also evaluated the HIBER dataset as a supplement and observed a significant improvement in performance. It should be noted that constraint is not always accurate when the subject jumps or stands on an obstacle, but it is still effective in terms of stabilizing inference.

\paragraph{2D vs. 3D Supervision Strength:}
Table~\ref{tab:loss_weight} compares the various weight parameters $\lambda_{\OP{3D}}$ and $\lambda_{\OP{2D}}$ in \eqref{eq:loss}.  The results highlight the necessity of accounting for the loss of both the 3D BBox and the 2D BBox, and the importance of the prediction accuracy of the 3D BBox, especially in the radar coordinate system, for the prediction accuracy of the 2D BBox on the image plane. The image plane supervision is essential to train the learnable refinement module. Strong 2D (image plane) and 3D (radar space) supervision yields better performance. 

\paragraph{Number of BBoxes in Training:}
We evaluate the impact of $N_{\OP{train}}$, the number of BBoxes for REXO and the number of queries for RETR, on the three AP metrics in Table~\ref{tab:num_bboxes_training}. It is seen that  $\OP{AP}$ tends to decrease as $N_{\OP{train}}$ increases for both methods, but this may be due to the number of BBoxes being too large relative to the number of subjects, since the maximum number of subjects in the MMVR dataset is three per frame. 

\paragraph{Dynamic Number of BBoxes in Inference:}
Table~\ref{fig:comp_n_bbox_inference} evaluates the impact of varying the number of BBoxes during inference. While RETR exhibits a sharp performance decline when the number of queries exceeds $10$, REXO experiences a much smaller decrease. This robustness in handling varying numbers of BBoxes during inference is a direct advantage inherited from DiffusionDet. 

\paragraph{Number of Iteration Steps:}
Table~\ref{fig:comp_diff_steps} presents REXO's performance as the number of iteration steps increases. Increasing the steps from $1$ to $10$ yields improvements of $+0.78$ in $\OP{AP}$, showing consistent gains with more iterations. We also report runtime and FPS on a single NVIDIA RTX 6000: 60 ms (17 FPS) for 1 step, 255 ms (4 FPS) for 5 steps, and 483 ms (2 FPS) for 10 steps. While more steps improve accuracy, they incur higher latency. Thus, using 5 steps offers a practical trade-off between detection performance and runtime efficiency for indoor human sensing.

\paragraph{Additional comparison with DiffusionDet and REXO:}
Table~\ref{tab:diffusiondet} confirms that our REXO (both views) outperforms the original DiffusionDet with radar heatmaps, which denotes our REXO is more appropriate for our radar settings. It also suggests the horizontal view (i.e., bird's-eye view) is more critical than the vertical view for detection since the vertical view cannot separate the azimuth position.

\subsection{Challenging Cases }
To better understand challenging configurations, we provide additional analysis for two scenarios: 1) when two subjects are at a similar depth; and 2) generalization over unseen environments. For 1), Fig.~\ref{fig:same_depth_heatmap} compares vertical heatmaps when a single subject (left) and two subjects (right) present at nearly the same depth but at different heights. 
When multiple subjects are at the same depth, the reflections from different subjects overlap and form more complex patterns than that for a single subject, potentially leading to failed cross-view feature association and radar-conditioned denoising steps. As confirmed in Table~\ref{tab:ap_drop_same_depth}, $\OP{AP}$ drops significantly when the depth difference $D$ is less than $20$ cm based on the evaluation over P2S2 of MMVR. For 2), we divide the test radar frames in P2S2 of MMVR into ``Seen" and ``Unseen" frames, and analyze their APs using IoU histograms in Fig.~\ref{fig:iou_histograms_p2s2}.  For the ``Seen" frames, REXO exhibits a histogram similar to one observed in P1S2. In contrast, for the ``Unseen'' frames, REXO clearly dominates the IoU range of $[0.75, 1]$, while RETR shows a heavier concentration around an IoU of $0.5$.

Fig.~\ref{fig:visualization_main} further visualizes selected ``Unseen" frames from a room never encountered during training in P2S2. It is seen that 2D BBox predictions by REXO align more closely with human segmentation masks (purple pixels) than those of RETR and RFMask. This improvement is potentially due to the explicit cross-view feature association, which strengthens consistency across radar views even in new environments, yielding better generalization.  More visualization examples are provided in Appendix~\ref{sec:more_ablation}.

\section{Conclusion}
\label{sec:conclusion}
For indoor radar perception, we proposed REXO, a novel multi-view radar object detection method that refines 3D BBoxes through a diffusion process. By explicitly guiding cross-view radar feature association and incorporating ground-level constraint, REXO achieves consistent performance improvements on two open indoor radar datasets over a list of strong baselines. 

\bibliography{aaai2026}

\clearpage
\appendix
\setcounter{secnumdepth}{2}

\section{Details on Diffusion Models}
\label{sec:detailedDiffusion}

Denoising Diffusion Probabilistic Models (DDPM)~\cite{Ho2020_DDPM} and Denoising Diffusion Implicit Models (DDIM)~\cite{Song2021_DDIM} are latent variable models designed to approximate the data distribution $q(\B{x}_0)$ using a generative model distribution $p_\theta(\B{x}_0)$. The generative model structure is expressed as:
\begin{align}
    p_\theta(\B{x}_0) &= \int p_\theta(\B{x}_{0:T}) d\B{x}_{1:T}, \\
    p_\theta(\B{x}_{0:T}) &= p_\theta(\B{x}_T) \prod_{t=1}^T p_\theta^{(t)}(\B{x}_{t-1} | \B{x}_t),
\end{align}
where $\B{x}_{1:T}$ are latent variables defined in the same space as $\B{x}_0$. The generative process reverses the forward diffusion process, transforming noise $\B{x}_T$ into $\B{x}_0$ over $T$ steps. Both models share a training objective based on maximizing the variational lower bound (ELBO):
\begin{align}
    &\max_\theta \mathbb{E}_{q(\B{x}_0)} [\log p_\theta(\B{x}_0)] \nonumber\\
    &\leq \max_\theta \mathbb{E}_{q(\B{x}_{0:T})} [\log p_\theta(\B{x}_{0:T}) - \log q(\B{x}_{1:T} | \B{x}_0)].
\end{align}
In both cases, the forward process $q(\B{x}_{1:T} | \B{x}_0)$ has no learnable parameter, simplifying training to focus on the generative process.

\paragraph{DDPM~\cite{Ho2020_DDPM}:}
The forward process progressively adds Gaussian noise to $\B{x}_0$ through a Markov chain:
\begin{equation}
    q(\B{x}_t | \B{x}_{t-1}) = \C{N} \left( \sqrt{\frac{\alpha_t}{\alpha_{t-1}}} \B{x}_{t-1}, \left(1 - \frac{\alpha_t}{\alpha_{t-1}}\right) \B{I} \right),
\end{equation}
where $\alpha_t$ is a decreasing sequence. This transforms $\B{x}_0$ into nearly pure Gaussian noise $\B{x}_T$, leading to a direct sampling of the $\B{x}_t$ from $\B{x}_0$:
\begin{align}
    \label{eq:noising_}
    q\LS{\B{x}_t \mid \B{x}_0} &=\C{N}\LS{\B{x}_t ; \sqrt{\bar{\alpha}_t} \B{x}_0,\LS{1-\bar{\alpha}_t} \B{I}},
\end{align}
where $\bar{\alpha}_{t} := \prod_{s=0}^t \alpha_s=\prod_{s=0}^t\LS{1-\beta_s}$ with $\beta_s$ denoting the noise variance schedule. The reverse process approximates this process backward:
\begin{equation}
    p_\theta^{(t)}(\B{x}_{t-1} | \B{x}_t) = \C{N} \left( \mu_\theta^{(t)}(\B{x}_t), \sigma_t^2 \B{I} \right),
\end{equation}
where $\mu_\theta^{(t)}(\B{x}_t) = \frac{1}{\sqrt{\alpha_t}}\left(\B{x}_t-\frac{\beta_t}{\sqrt{1-\bar{\alpha}_t}} \B{\epsilon}_{\theta}^{\LS{t}}\left(\B{x}_t\right)\right)$ is a learnable function and $\B{\epsilon}_\theta^{(t)}$ is designed with a neural network model with learnable parameter $\theta$. The sampling can be done following \eqref{eq:sampling}. The loss function is designed to minimize the discrepancy in predicting noise:
\begin{equation}
    L_\gamma(\B{\epsilon}_\theta) = \sum_{t=1}^T \gamma_t \mathbb{E}_{\B{x}_0, \B{\epsilon}} \left[ \|\B{\epsilon}_\theta^{\LS{t}}(\B{x}_t) - \B{\epsilon}\|_2^2 \right],
\end{equation}
where $\B{\epsilon} \sim \C{N}(\B{0}, \B{I})$ is Gaussian noise, and $\gamma_t$ are weights.

\paragraph{DDIM~\cite{Song2021_DDIM}:}
DDIM generalizes the forward process to a non-Markovian framework which can be derived from Bayes’ rule:
\begin{equation}
    q\left(\B{x}_t\mid\B{x}_{t-1}, \B{x}_0\right) = \frac{q\left(\B{x}_{t-1} \mid \B{x}_t, \B{x}_0\right) q\left(\B{x}_t \mid \B{x}_0\right)}{q\left(\B{x}_{t-1} \mid \B{x}_0\right)},
\end{equation}
where $q\left(\B{x}_t \mid \B{x}_0\right)$ and $q\left(\B{x}_{t-1} \mid \B{x}_0\right)$ are defined with \eqref{eq:noising_} and
\begin{align}
    &q(\B{x}_{t-1} | \B{x}_t, \B{x}_0) \nonumber\\
    &= \C{N} \left( \sqrt{\alpha_{t-1}} \B{x}_0 + \sqrt{1 - \alpha_{t-1} - \sigma_t^2} \B{\epsilon}^{\LS{t}}, \sigma_t^2 \B{I} \right),
\end{align}
with
\begin{align}
    \B{\epsilon}^{\LS{t}} = \frac{\B{x}_t - \sqrt{\alpha_t} \B{x}_0}{\sqrt{1 - \alpha_t}}.
\end{align}
Different from DDPM, this process maintains a direct dependence on the original data $\B{x}_0$ at each step.
The reverse process in DDIM reconstructs $\B{x}_0$ as 
\begin{align}
    \B{x}_{t-1} &= \sqrt{\alpha_{t-1}} \hat{\B{x}}_{0}  + \sqrt{1-\alpha_{t-1}-\sigma_t^2} \cdot \B{\epsilon}^{t} + \sigma_t \B{\epsilon}_t,\\
    \B{\epsilon}^{\LS{t}} &= \frac{\B{x}_t - \sqrt{\alpha_t} \hat{\B{x}}_0}{\sqrt{1 - \alpha_t}},
\end{align}
where $\B{\epsilon}^{\LS{t}}$ is the estimated noise from $\B{x}_0$ to $\B{x}_t$.
The loss function remains identical to that in DDPM which uses the neural network $\B{\epsilon}_\theta^{\LS{t}}(\B{x}_t)$, ensuring the compatibility in training.

DDIM enables a faster sampling by reducing the number of steps $T$ in the sampling process. A subset of latent variables $\{\B{x}_{\tau_1}, \dots, \B{x}_{\tau_S}\}$ is defined where $\tau$ is an increasing sub-sequence of $\LL{1,\cdots,T}$ of length $S$, and sampling occurs over this shortened trajectory:
\begin{align}
    \B{x}_{\tau_{i-1}} &= \sqrt{\alpha_{\tau_{i-1}}} \hat{\B{x}}_{0} + \sqrt{1-\alpha_{\tau_{i-1}}-\sigma_{\tau_i}^2} \cdot \B{\epsilon}^{\LS{{\tau_i}}} + \sigma_{\tau_i} \B{\epsilon}_{\tau_i},\nonumber\\
    \B{\epsilon}^{\LS{{\tau_i}}} &= \frac{\B{x}_{\tau_i} - \sqrt{\alpha_{\tau_i}} \hat{\B{x}}_{0}}{\sqrt{1-\alpha_{\tau_i}}}.
\end{align}
For instance, when $S \ll T$, computational efficiency is significantly improved.
This method allows a pre-trained DDPM model to be reused while achieving $10\times$ to $50\times$ faster sampling. This acceleration is particularly beneficial for scenarios requiring low-latency processing such as perception tasks.

For REXO training in Fig.~\ref{fig:architecture_train} and inference, we construct a $\OP{DenoisingDet}_{\theta}$ that directly predicts $\B{x}_0$, similar to the above DDIM and DiffusionDet ~\cite{Chen2023_diffusiondet}. This direct estimation of $\B{x}_0$ allows us to compute the set-prediction loss in DETR \cite{Carion2020_detr} over both radar and camera coordinate systems. As a result, the noise prediction network $\B{\epsilon}^{\LS{\tau_i}}$ is a function of the predicted $\hat{\B{x}}_0$.

\section{Details of Cross-View Radar-Conditioned Denoising Detector}
\label{sec:denoisingDet}

\begin{figure}
    \centering
    \includegraphics[width=\linewidth]{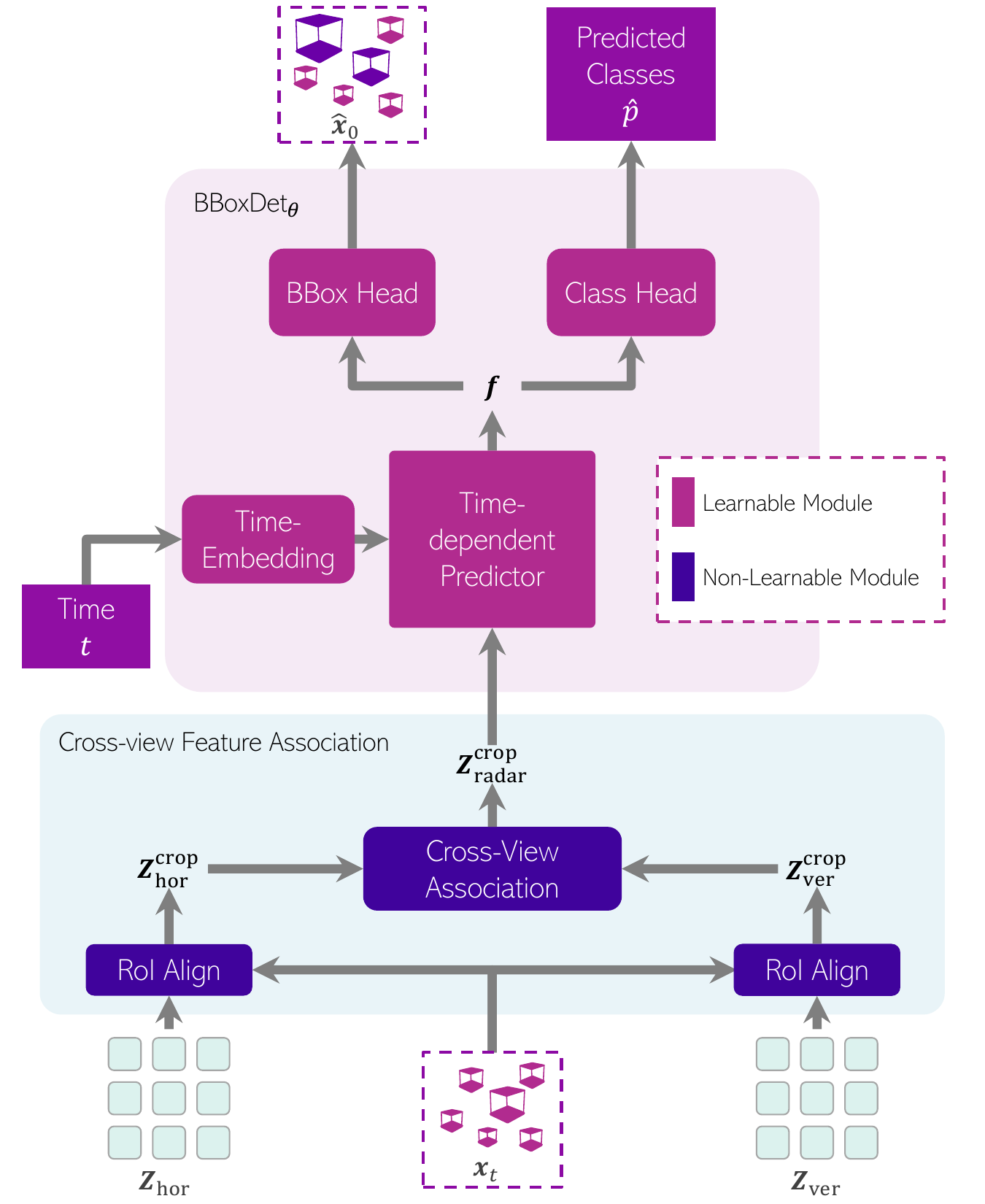}
    \caption{Overall architecture of the cross-view radar-conditioned denoising detector $\OP{DenoisingDet}_{\theta}$.  
    }
    \label{fig:association}
\end{figure}

Fig.~\ref{fig:association} illustrates the overall architecture of the proposed cross-view radar-conditioned denoising detector
\begin{align}
    \hat{\B{x}}_0 = \OP{DenoisingDet}_{\theta}\left( \B{x}_{t}, t, \B{Z}_{\OP{hor}},\B{Z}_{\OP{ver}}\right),
\end{align}
which takes as input the noisy 3D BBox set $\B{x}_t$, the time index $t$, and the two radar view backbone features $\{\B{Z}_{\OP{hor}},\B{Z}_{\OP{ver}} \}$. The output is a direct estimation of $\B{x}_0$. Specifically, $\OP{DenoisingDet}_{\theta}$ comprises two key modules: 
\begin{itemize}
\item \textbf{1. Cross-view feature association};
\item \textbf{2. BBox detector $\OP{BBoxDet}_{\theta}$}. 
\end{itemize}

The cross-view feature association module was explained in~\eqref{eq:BBox_projection}-\eqref{eq:feature_association}, which involves the 3D-to-2D BBox projection of \eqref{eq:BBox_projection}, feature cropping via the RoIAlign operation of \eqref{eq:roialign}, and the feature aggregation of \eqref{eq:feature_association}, yielding $\B{Z}_{\OP{radar}}^{\OP{crop}} = \LM{\B{Z}_{\OP{hor}}^{\OP{crop}}, \B{Z}_{\OP{ver}}^{\OP{crop}}} \in\R^{ C \times r \times 2r}$ in Fig.~\ref{fig:association} (blue hatch). 

The BBox detector module, $\OP{BBoxDet}_{\theta}$, first incorporates a time-dependent predictor, $\OP{Predictor}$, designed to extract time-dependent features $\B{f}$ from the aligned cross-view radar features $\B{Z}_{\OP{radar}}^{\OP{crop}}$ and the time-embedding of $t$~\cite{Ho2020_DDPM},
\begin{equation}
    \B{f} = \OP{Predictor}\LS{\OP{e}_{t}, \B{Z}_{\OP{radar}}^{\OP{crop}}},
\end{equation}
where $\OP{e}_{t} = \OP{Embedding}\LS{t}$ is a time-embedding of time $t$.
In the $\OP{Predictor}$, self-attention is applied to the cross-view radar features $\B{Z}_{\OP{radar}}^{\OP{crop}}$ to reason about the relations between objects. Then, the features are enhanced by applying the dynamic convolution to each 3D RoI feature. Finally, they are fed into the diffusion process by time-embedding to obtain the feature $\B{f}$.

Using the time-dependent feature $\B{f}$, the BBox and Class heads are used to predict the BBoxes $\hat{\B{x}}_0$ and the class labels. 
In the BBox head, the offset $\Delta{\B{x}}$ is first computed as 
\begin{align}
    \Delta{\B{x}} &= \LM{\Delta{x}, \Delta{y}, \Delta{z}, \Delta{w}, \Delta{h}, \Delta{d}}^{\top}\in\R^{6}\nonumber\\
    &= \OP{FFN}_{\OP{offset}}^{\OP{3D}}\LS{\B{f}},
\end{align}
where $\OP{FFN}_{\OP{offset}}^{\OP{3D}}:\R^{D}\to\R^{6}$ represents a feed-forward network (FFN) to estimate the 3D BBox offsets and each $\Delta*$ denotes the offset for the corresponding BBox parameter ($x, y, z, w, h, d$). To ensure numerical stability, the values of $\Delta{w}$, $\Delta{h}$ and $\Delta{d}$ are clipped to a maximum of $\log\LS{10^5 / 16}$. The BBox estimate $\hat{\B{x}}_{0}$ is then computed using a linear update for the BBox center position and an exponential scaling over the axis length offsets $\LM{e^{\Delta{w}},e^{\Delta{h}},e^{\Delta{d}}}$:
\begin{equation}
    \label{eq:pred_3d_bbox}
    \hat{\B{x}}_{0} \!=\! \LM{c_x\!+\!w\Delta{x},c_y\!+\!h\Delta{y},c_{z}\!+\!d\Delta{z},e^{\Delta{w}}w,e^{\Delta{h}}h,e^{\Delta{d}}d}^{\top}.
\end{equation}

In the class head, the class scores can be predicted as $\hat{\B{p}}$. The classification loss evaluates the alignment between the predicted objects and GT classes, serving as a crucial component in object detection tasks. The classification loss is defined as:
\begin{equation}
    \C{L}_{\OP{class}} = -\sum_{i=1}^{N_{\OP{train}}} {\log{\hat{p}_{\hat{\sigma}\LS{i}}\LS{c_{i}}}},
\end{equation}
where $c_i$ is the GT class label, $\hat{p}_{\hat{\sigma}(i)}(c_i)$ denotes the predicted probability for class $c_i$, and $\hat{\sigma}(i)$ represents the optimal matching computed by the Hungarian algorithm. It is a permutation that defines the one-to-one correspondence between the GT object set and the predicted set. Additionally, when the GT class is $\emptyset$ (no object), the loss term is down-weighted to address class imbalance. This design improves the model's performance in predicting positive examples (objects present in the image).

\section{Details of 3D-to-2D Projection and Necessity of the Refinement Module}
\label{sec:details_refinement_module}
We present the detailed explanations for 3D-to-2D projection and necessity of the refinement module.
Given a 3D BBox which consists of its eight vertices
\begin{equation}
    \{\B{x}_{\OP{camera}}^i \in \mathbb{R}^3 \mid i = 1, \dots, 8\},
\end{equation}
where each $\B{x}_{\OP{camera}}^i$ is expressed in the 3D camera coordinate system, our goal is to compute the corresponding 2D BBox $\B{b}_{\OP{init}} \in \R^4$, defined by its center coordinates $(x_c, y_c)$ and its width $w$ and height $h$.
To achieve this, we define a projection function with a pinhole camera model as a concrete expression of \eqref{eq:image_plane_projection}:
\begin{equation}
    \OP{proj}_{\OP{pinhole}}: \mathbb{R}^3 \to \mathbb{R}^2: \LS{X,Y,Z}\mapsto\LS{p_x,p_y}.
\end{equation}
In this model, the projection of the point $\B{x}_{\OP{camera}}^* = \LS{X,Y,Z}$ onto the image plane is given by
\begin{equation}
    p_x = \frac{f_x X}{Z} + c_x, \quad p_y = \frac{f_y Y}{Z} + c_y,
\end{equation}
where $f_x$ and $f_y$ are the focal lengths along the $x$ and $y$ axes (in pixels), and $(c_x, c_y)$ represents the coordinates of the principal point in the image. In homogeneous coordinates, this mapping can be expressed as
\begin{equation}
\lambda \begin{pmatrix} p_x \\ p_y \\ 1 \end{pmatrix} = \begin{pmatrix} f_x & 0 & c_x \\ 0 & f_y & c_y \\ 0 & 0 & 1 \end{pmatrix} \begin{pmatrix} X \\ Y \\ Z \end{pmatrix},
\end{equation}
with the scaling factor $\lambda = Z$. 
Thus, for each vertex, the projection onto the image plane is given by:
\begin{equation}
    \B{p}^i = \OP{proj}_{\OP{pinhole}}(\B{x}_{\OP{camera}}^i), \quad \text{for } i = 1, \dots, 8,
\end{equation}
where $\B{p}^i = (p_x^i, p_y^i)$ represents the 2D coordinates of the projected point in the image plane.
Once the eight vertices have been projected, the extreme coordinates on the image plane are determined as:
\begin{equation}
    u_{\min} = \min_{i} \{ p_x^i \}, \quad u_{\max} = \max_{i} \{ p_x^i \},
\end{equation}
\begin{equation}
v_{\min} = \min_{i} \{ p_y^i \}, \quad v_{\max} = \max_{i} \{ p_y^i \}.
\end{equation}
Using these extremes, the center coordinates, width, and height of the 2D BBox are computed by:
\begin{equation}
    x_c = \frac{u_{\min} + u_{\max}}{2}, \quad y_c = \frac{v_{\min} + v_{\max}}{2},
\end{equation}
\begin{equation}
    w = u_{\max} - u_{\min}, \quad h = v_{\max} - v_{\min}.
\end{equation}
Thus, the final 2D BBox can be obtained as:
\begin{equation}
    \B{b}_{\OP{init}} = \left( x_c, y_c, w, h \right).
\end{equation}

\begin{figure}
    \centering
    \includegraphics[width=0.9\linewidth]{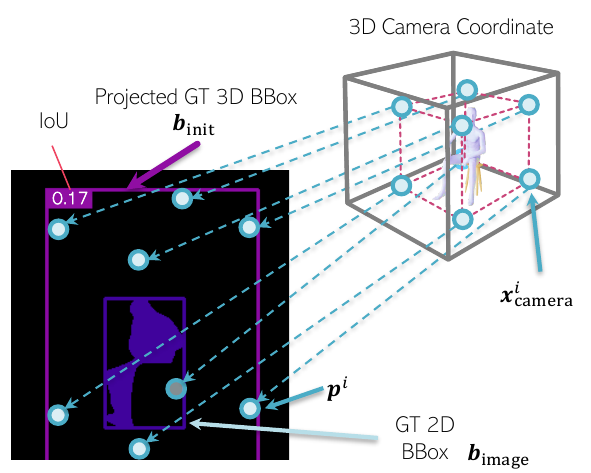}
    \caption{A direct projection of 3D BBoxes to the 2D image plane results in oversized 2D BBoxes. A learnable module is used to refine the projected BBoxes close to the 2D BBox GT.}
    \label{fig:2d_supervision}
\end{figure}

\begin{figure}
    \centering
    \includegraphics[width=0.6\linewidth]{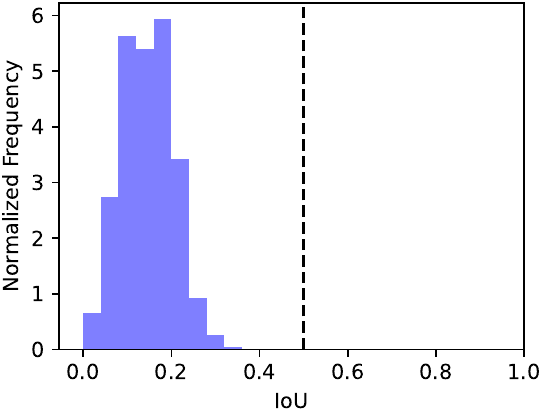}
    \caption{IoU histogram when no image plane supervision. Almost all IoU values are lower than 0.5, resulting in 0 AP.}
    \label{fig:iou_histograms_gt}       
\end{figure}

The 2D BBoxes obtained by projection, as shown by the purple box $\B{b}_{\OP{init}}$ in Fig.~\ref{fig:2d_supervision}, are often too large. This occurs because projecting the eight vertices $\B{x}_{\OP{camera}}^i$ captures the depth information from the camera, which causes both the near and far parts of the object to be displayed in a 3D manner. As a result, to accurately predict the 2D BBox $\B{b}_{\OP{image}}$ on the image plane, we must use a refinement module. This module reduces the size of the initial BBox, as illustrated by the blue boxes in Fig.~\ref{fig:2d_supervision}.

To better understand the need for refinement, we calculated the Intersection over Union (IoU) between the ground-truth (GT) 3D BBoxes (projected from the 3D space) and the GT 2D BBoxes (defined on the image plane). The histogram of IoU values in Fig.~\ref{fig:iou_histograms_gt} shows a roughly Gaussian distribution with a peak around $0.15$, and nearly all IoU values are below $0.5$. In fact, in Fig.~\ref{fig:2d_supervision}, the IoU is $0.17$. This indicates that if we do not apply refinement, even when the 3D BBoxes are correctly predicted in the radar coordinate system, the average precision (AP) on the image plane would be zero. Therefore, our REXO method uses a refinement module.

\section{RFMask3D as a Baseline}
\label{sec:rfmask3D}

\begin{figure}
    \centering
    \includegraphics[width=0.9\linewidth]{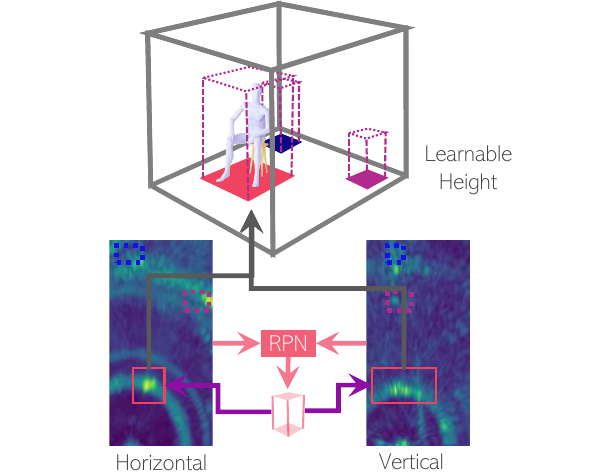}
    \caption{3D Proposals with RFMask3D.}
    \label{fig:rfmask3d}
    \vspace{-5mm}
\end{figure}
\begin{table*}[t]
    \centering
    \footnotesize
    \setlength\tabcolsep{9.5pt}
    \caption{Details of hyper-parameters. Fixed height for the HIBER dataset depends on the environment.}
    \begin{tabular}{clcccccc}
        \toprule
        \multicolumn{2}{c}{\multirow{2}{*}{\textbf{Name}}} & \multirow{2}{*}{\textbf{Notation}} & \multicolumn{4}{c}{\textbf{Value}} \\
        \cline{4-7}
         & & & \rule{0pt}{10pt}P1S1 & P1S2 & P2S1 & P2S2 \\
        \midrule
        \multirow{9}{*}{\rotatebox{90}{\textbf{Data}}}
         & \# of training & - & 86579 & 70266 & 190441 & 118280 \\
         & \# of validation & - & 10538 & 24398 & 23899 & 33841 \\
         & \# of test & - & 10785 & 13238 & 23458 & 85677 \\
         & Input radar heatmap size & $H\times W$ & 256$\times$128 & 256$\times$128 & 256$\times$128 & 256$\times$128 \\
         & Segmentation mask size & $H\times W$ & 240$\times$320 & 240$\times$320 & 240$\times$320 & 240$\times$320 \\
         & Resolution of range & cm & 11.5 & 11.5 & 11.5 & 11.5\\
         & Resolution of azimuth & deg. &  1.3 & 1.3 & 1.3 &  1.3 \\
         & Resolution of elevation & deg. &  1.3 &  1.3 &  1.3 &  1.3\\
         & Scale & - & log & log & log & log\\
        \midrule
        \multirow{10}{*}{\rotatebox{90}{\textbf{Model}}}
         & Backbone & - & ResNet18 & ResNet18 & ResNet18 & ResNet18 \\
         & \# of input consecutive radar frames & - & 4 & 4 & 4 & 4 \\
         & Extracted feature map size & $H/s \times W/s$ & 64$\times$32 & 64$\times$32 & 64$\times$32 & 64$\times$32 \\
         & Threshold for detection & - & 0.5 & 0.5 & 0.5 & 0.5 \\
         & Loss weight for $\OP{GIoU}$ on radar coordinate system & $\lambda_{\OP{GIoU}}$ & 2.0 & 2.0 & 2.0 & 2.0 \\
         & Loss weight for $\OP{GIoU}$ on image plane & $\lambda_{\OP{GIoU}}$ & 2.0 & 2.0 & 2.0 & 2.0 \\
         & Loss weight for $\OP{L_1}$ on radar coordinate system & $\lambda_{\OP{L_1}}$ & 5.0 & 5.0 & 5.0 & 5.0 \\
         & Loss weight for $\OP{L_1}$ on image plane & $\lambda_{\OP{L_1}}$ & 5.0 & 5.0 & 5.0 & 5.0 \\
         & Loss weight for $\OP{radar}$ & $\lambda_{\OP{3D}}$ & 1.0 & 1.0 & 1.0 & 1.0 \\
         & Loss weight for $\OP{image}$ & $\lambda_{\OP{2D}}$ & 1.0 & 1.0 & 1.0 & 1.0 \\
        \midrule
        \multirow{13}{*}{\rotatebox{90}{\textbf{Training}}} 
         & Batch size & - & 32 & 32 & 32 & 32 \\
         & Epoch for detection & - & 100 & 100 & 100 & 100\\
         & Patience for early stopping & - & 5 & 5  & 5 & 5\\
         & Check val every $n$ epoch for early stopping  & - & 2 & 2 & 2 & 2 \\
         & Optimizer & - &  AdamW & AdamW & AdamW & AdamW \\
         & Learning rate & - & 1e-4 & 1e-4 & 1e-4 & 1e-4 \\
         & Sheduler & - & Cosine & Cosine & Cosine & Cosine \\
         & Maximum number of epochs for sheduler & - & 100 & 100 & 100 & 100 \\
         & Weight decay & - & 1e-3 & 1e-3 & 1e-3 & 1e-3 \\
         & \# of workers & - & 8 & 8 & 8  & 8  \\
         & GPU (NVIDIA) & - & A40 & A40 & A40 & A40  \\
         & \# of GPUs & - & 1 & 1 & 1 & 1  \\
         & Approximate training time & day & 1 & 1 & 2 & 2 \\
        \bottomrule
    \end{tabular}
    \label{tab:detailed_hyperparameters}
\end{table*}
As one of the baselines in our evaluation experiments, we constructed RFMask3D by extending RFMask~\cite{Wu2023_RFMask} to 3D. RFMask uses a region proposal network (RPN) to extract regions of interest (RoIs) from a horizontal heatmap based on 2D anchor boxes and predicts 3D BBoxes in the 3D radar coordinate system by combining them with fixed heights. By designing an RPN that uses 3D anchor boxes, we explicitly extract 3D RoIs from both horizontal and vertical heatmaps, as shown in Fig.~\ref{fig:rfmask3d}, enabling the estimation of 3D BBoxes. Unlike RFMask, this method allows for the learning of height as well.

\section{Hyperparameters for Performance Evaluation}
\label{sec:experimental_setting}

The hyper-parameters used in our experiments of Section~\ref{sec:experiments} are shown in Table~\ref{tab:detailed_hyperparameters}. The table is divided into three parts, Data, Model, and Training, each with parameter names, notations, and values for each dataset.
\begin{table}[t]
    \footnotesize
    \centering
    \caption{Full Result of Table~\ref{tab:main_mmvr} under P1S1 on MMVR dataset.}
    \setlength\tabcolsep{8pt}
    \begin{tabular}{l| rrrrr}
        \toprule
        \multicolumn{1}{c}{$\OP{Method}$} & \multicolumn{1}{c}{$\OP{AP}$} & \multicolumn{1}{c}{$\OP{AP_{50}}$} & \multicolumn{1}{c}{$\OP{AP_{75}}$} & \multicolumn{1}{c}{$\OP{AR_{1}}$} & \multicolumn{1}{c}{$\OP{AR_{10}}$} \\
        \midrule
        RFMask & 25.53 & 67.30 & 15.86 & 38.90 & 38.91 \\
        RFMask3D & 34.84 & 69.57 & 31.74 & 44.88 & 44.88 \\
        DETR & 35.64 & 77.59 & 28.00 & - & - \\
        RETR & \cellcolor{gray!20}39.62 & \cellcolor{gray!20}80.55 & 33.84 & \cellcolor{gray!20}51.90 & \cellcolor{gray!20}53.42 \\
        \midrule
        REXO & 39.23 & 73.46 & \cellcolor{gray!20}37.83 & 48.58 & 48.58 \\
        \bottomrule
    \end{tabular}
    \label{tab:full_main_p1s1}
\end{table}

\begin{table}[t]
    \footnotesize
    \centering
    \caption{Full Result of Table~\ref{tab:main_mmvr} under P1S2 on MMVR dataset.}
    \setlength\tabcolsep{8pt}
    \begin{tabular}{l| rrrrr}
        \toprule
        \multicolumn{1}{c}{$\OP{Method}$} & \multicolumn{1}{c}{$\OP{AP}$} & \multicolumn{1}{c}{$\OP{AP_{50}}$} & \multicolumn{1}{c}{$\OP{AP_{75}}$} & \multicolumn{1}{c}{$\OP{AR_{1}}$} & \multicolumn{1}{c}{$\OP{AR_{10}}$} \\
        \midrule
        RFMask & 24.46 & 66.82 & 11.22 & 34.07 & 34.53 \\
        RFMask3D & 30.75 & 76.48 & 16.23 & 40.28 & 40.29 \\
        DETR & 28.51 & 75.90 & 13.42 & - & - \\
        RETR & 30.16 & 78.95 & 15.17 & 42.93 & 42.93 \\
        \midrule
        REXO & \cellcolor{gray!20}36.48 & \cellcolor{gray!20}87.02 & \cellcolor{gray!20}20.51 & \cellcolor{gray!20}47.70 & \cellcolor{gray!20}47.71 \\
        \bottomrule
    \end{tabular}
    \label{tab:full_main_p1s2}
\end{table}

\begin{table}[t]
    \footnotesize
    \centering
    \caption{Full Result of Table~\ref{tab:main_mmvr} under P2S1 on MMVR dataset.}
    \setlength\tabcolsep{8pt}
    \begin{tabular}{l| rrrrr}
        \toprule
        \multicolumn{1}{c}{$\OP{Method}$} & \multicolumn{1}{c}{$\OP{AP}$} & \multicolumn{1}{c}{$\OP{AP_{50}}$} & \multicolumn{1}{c}{$\OP{AP_{75}}$} & \multicolumn{1}{c}{$\OP{AR_{1}}$} & \multicolumn{1}{c}{$\OP{AR_{10}}$} \\
        \midrule
        RFMask & 31.37 & 61.50 & 27.48 & 33.07 & 38.21 \\
        RFMask3D & 39.89 & 80.38 & 35.35 & 36.81 & 48.54 \\
        DETR & 29.53 & 63.08 & 25.35 & - & - \\
        RETR & 46.75 & 83.80 & 46.06 & 42.19 & 57.39 \\
        \midrule
        REXO & \cellcolor{gray!20}48.35 & \cellcolor{gray!20}85.89 & \cellcolor{gray!20}48.38 & \cellcolor{gray!20}43.52 & \cellcolor{gray!20}57.88 \\
        \bottomrule
    \end{tabular}
    \label{tab:full_main_p2s1}
\end{table}

\begin{table}[t]
    \footnotesize
    \centering
    \caption{Full Result of Table~\ref{tab:main_mmvr} under P2S2 on MMVR dataset.}
    \setlength\tabcolsep{8pt}
    \begin{tabular}{l| rrrrr}
        \toprule
        \multicolumn{1}{c}{$\OP{Method}$} & \multicolumn{1}{c}{$\OP{AP}$} & \multicolumn{1}{c}{$\OP{AP_{50}}$} & \multicolumn{1}{c}{$\OP{AP_{75}}$} & \multicolumn{1}{c}{$\OP{AR_{1}}$} & \multicolumn{1}{c}{$\OP{AR_{10}}$} \\
        \midrule
        RFMask & 6.03 & 22.77 & 0.88 & 9.25 & 12.09 \\
        RFMask3D & 12.26 & 37.01 & 4.34 & 18.91 & 19.52 \\
        DETR & 9.29 & 34.69 & 2.49 & 20.68 & 22.82 \\
        RETR & 12.45 & 41.30 & 4.96 & 19.96 & 21.58 \\
        \midrule
        REXO & \cellcolor{gray!20}23.47 & \cellcolor{gray!20}64.41 & \cellcolor{gray!20}10.44 & \cellcolor{gray!20}30.65 & \cellcolor{gray!20}33.44 \\
        \bottomrule
    \end{tabular}
    \label{tab:full_main_p2s2}
\end{table}

\begin{table}[t]
    \footnotesize
    \centering
    \caption{Full Result of Table~\ref{tab:main_hiber} under WALK on HIBER dataset.}
    \setlength\tabcolsep{8pt}
    \begin{tabular}{l| rrrrr}
        \toprule
        \multicolumn{1}{c}{$\OP{Method}$} & \multicolumn{1}{c}{$\OP{AP}$} & \multicolumn{1}{c}{$\OP{AP_{50}}$} & \multicolumn{1}{c}{$\OP{AP_{75}}$} & \multicolumn{1}{c}{$\OP{AR_{1}}$} & \multicolumn{1}{c}{$\OP{AR_{10}}$} \\
        \midrule
        RFMask & 17.77 & 52.46 & 6.78 & 32.71 & 32.71 \\
        RFMask3D & 16.58 & 48.10 & 6.53 & 29.89 & 29.89 \\
        DETR & 14.45 & 47.33 & 4.25 & 28.64 & 28.64 \\
        RETR & 22.09 & 59.83 & 10.99 & \cellcolor{gray!20}35.16 & 35.16 \\
        \midrule
        REXO & \cellcolor{gray!20}25.33 & \cellcolor{gray!20}62.55 & \cellcolor{gray!20}15.83 & 20.03 & \cellcolor{gray!20}37.54 \\
        \bottomrule
    \end{tabular}
    \label{tab:full_main_walk}
\end{table}

\begin{figure}[t]
    \centering
    \includegraphics[width=0.9\linewidth]{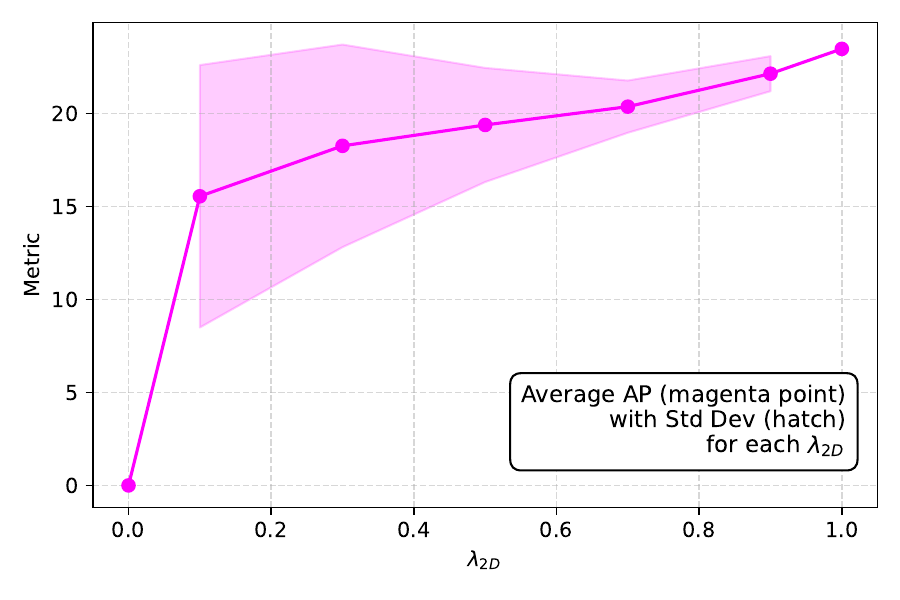}
    \caption{Result of Table~\ref{tab:loss_weight}: The effect of 2D image supervision with loss weight $\lambda_{\OP{2D}}$ while keeping $\lambda_{\OP{3D}}=1.0$.}
    \label{tab:full_loss_weight}
\end{figure}

\begin{table}[t]
    \footnotesize
    \centering
    \caption{Full Result of Table~\ref{tab:num_bboxes_training}: The number of BBoxes in Training.}
    \setlength\tabcolsep{6.1pt}
    \begin{tabular}{cc| rrrrr}
        \toprule
        \multicolumn{1}{c}{$\OP{Method}$} & $N_{\OP{train}}$ & \multicolumn{1}{c}{$\OP{AP}$} & \multicolumn{1}{c}{$\OP{AP_{50}}$} & \multicolumn{1}{c}{$\OP{AP_{75}}$} & \multicolumn{1}{c}{$\OP{AR_{1}}$} & \multicolumn{1}{c}{$\OP{AR_{10}}$} \\
        \midrule
        RETR & 10 & \cellcolor{gray!20}12.45 & \cellcolor{gray!20}41.30 & \cellcolor{gray!20}4.96 & \cellcolor{gray!20}19.96 & \cellcolor{gray!20}21.58 \\
        RETR & 20 & 9.85 & 31.01 & 4.48 & 17.90 & 18.72 \\
        RETR & 50 & 8.49 & 29.76 & 2.53 & 17.12 & 18.91 \\
        \midrule
        REXO & 10 & \cellcolor{gray!20}23.47 & 64.41 & \cellcolor{gray!20}10.44 & \cellcolor{gray!20}30.65 & \cellcolor{gray!20}33.44 \\
        REXO & 20 & 20.94 & \cellcolor{gray!20}65.03 & 5.90 & 27.45 & 29.63 \\
        REXO & 50 & 19.67 & 61.44 & 5.63 & 26.30 & 28.45 \\
        \bottomrule
    \end{tabular}
    \label{tab:full_num_bboxes_training}
\end{table}

\begin{figure}[t]
    \centering
    \includegraphics[width=0.9\linewidth]{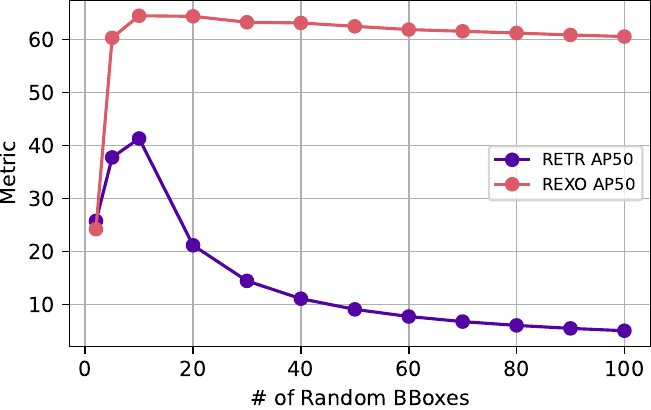}
    \caption{Full Result of Table~\ref{fig:comp_n_bbox_inference}: Dynamic number of BBoxes in inference.}
    \label{fig:full_comp_n_bbox_inference}
\end{figure}

\begin{figure}[t]
    \centering
    \includegraphics[width=0.9\linewidth]{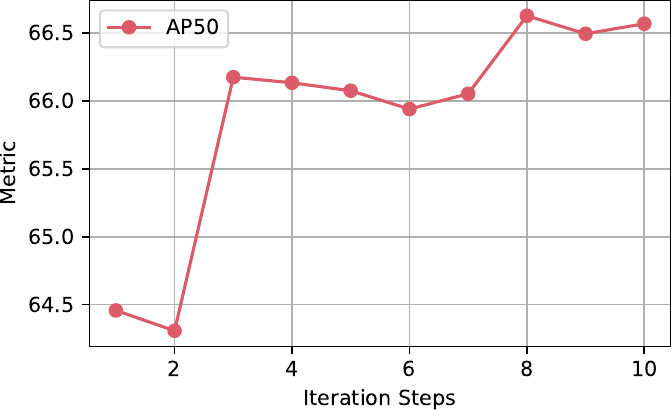}
    \caption{Full Result of Table~\ref{fig:comp_diff_steps}: \textbf{\# of denoising steps $S$}. More steps slightly improve the detection.}
    \label{fig:full_comp_diff_steps}
\end{figure}

\begin{table}[t]
    \centering
    \footnotesize
    \captionof{table}{Effect of denoising only one parameter.}
    \setlength\tabcolsep{6.5pt}
    \begin{tabular}{c|c|ccc}
        \toprule
        $\OP{Denoised\;Param.}$ & $\OP{\#\;of\;steps}$ & $\OP{AP}$ & $\OP{AP}_{50}$ & $\OP{AP}_{75}$ \\
        \midrule
        $c_z$ & 10 & 23.39 & 65.71 & 10.13 \\
        $d$   & 10  & 23.42 & 65.86 & 10.12\\
        All   & 10 & \cellcolor{gray!20}24.27 & \cellcolor{gray!20}66.57 & \cellcolor{gray!20}11.18 \\
        \bottomrule
    \end{tabular}
    \label{tab:factorization}
\end{table}

\begin{figure}[!htbp]
    \centering
    \includegraphics[width=0.9\linewidth]{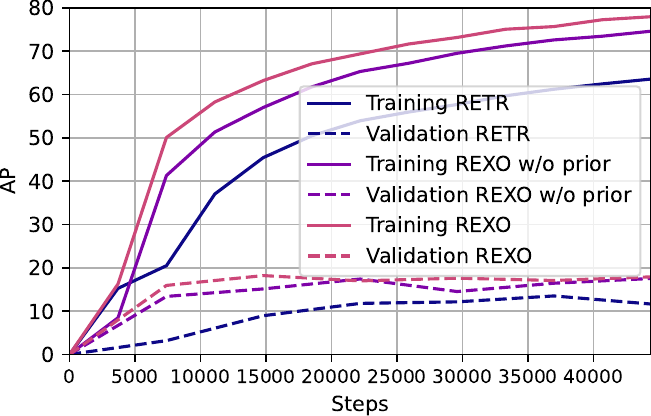}
    \caption{REXO can contribute to fast convergence by using a ground-level constraint for 3D BBox Diffusion.}
    \label{fig:full_comp_convergence}
    \vspace{-3mm}
\end{figure}
\section{Definition of Metrics}
\label{sec:definition_metrics}

\paragraph{Mean Intersection over Union:}
We adopt average precision on intersection over union~(IoU)~\cite{Everingham2010_pascalvoc} as an evaluation metric. 
IoU is the ratio of the overlap to the union of a predicted BBox $A$ and annotated BBox $B$ as:
\begin{equation}\label{eq:iou}
    \OP{IoU}\LS{A, B} = \frac{|A \bigcap B|}{|A \bigcup B|}.
\end{equation}

\paragraph{Average Precision:}
Average Precision (AP) can then be defined as the area under the interpolated precision-recall curve, which can be calculated using the following formula:
\begin{align}
    \OP{AP} &= \sum_{i=1}^{n-1}\LS{r_{i+1}-r_i} p_{\OP{interp}}\LS{r_{i+1}},\\
    p_{\OP{interp}}\LS{r} &= \max _{r^{\prime} \geq r} p\LS{r^{\prime}},
\end{align}
where the interpolated precision 
$p_{\OP{interp}}$ at a certain recall level $r$ is defined as the highest precision found for any recall level $r^{\prime} \geq r$.
We present three variants of average precision: $\OP{AP}_{50}$, $\OP{AP}_{75}$, and $\OP{AP}$, where the former two represent the loose and strict constraints of IoU, while $\OP{AP}$ is the averaged score over $10$ different IoU thresholds in $[0.5, 0.95]$ with a step size of $0.05$.

\paragraph{Average Recall:}
Average recall (AR)~\cite{Hosang2016_AverageRecall} between 0.5 and 1 of $\OP{IoU\; overlap\; threshold}$ can be computed by averaging over the overlaps of each annotation $\mathrm{gt}_{i}$ with the closest matched proposal, that is integrating over the $y:\OP{recall}$ axis of the plot instead of the $x:\OP{IoU\; overlap\; threshold}$ axis. Let $o$ be the $\OP{IoU}$ overlap and $\OP{recall}\LS{o}$ the function. Let $\OP{IoU}\LS{\mathrm{gt}_{i}}$ denote the $\OP{IoU}$ between the annotation $\mathrm{gt}_{i}$ and the closest detection proposal:
\begin{align}
\OP{AR} &= 2 \int_{0.5}^1 \OP{recall}(o) \mathrm{d} o\\
        &=\frac{2}{n} \sum_{i=1}^n \max \LS{\OP{IoU}\LS{\mathrm{gt}_i}-0.5,0}.
\end{align}
The following are some variations of $\OP{AR}$:
\begin{itemize}
    \item $\OP{AR}_{1}$: $\OP{AR}$ given one detection per frame;
    \item $\OP{AR}_{10}$: $\OP{AR}$ given 10 detection per frame;
    \item $\OP{AR}_{100}$: $\OP{AR}$ given 100 detection per frame.
\end{itemize}

\section{Ablation Study}
\label{sec:more_ablation}
\paragraph{Complete Results:}
Tables~\ref{tab:full_main_p1s1}, \ref{tab:full_main_p1s2}, \ref{tab:full_main_p2s1} and \ref{tab:full_main_p2s2} show the complete results of Table~\ref{tab:main_mmvr}, including $\OP{AR}$.
We added the metric regarding average recall explained in Appendix~\ref{sec:definition_metrics}.
The results show that the same trends as in Table~\ref{tab:main_mmvr} also apply to $\OP{AR}$. In particular, the generalization performance of Split P2S2 has improved significantly.

\paragraph{2D Image Supervision:}
Fig.~\ref{tab:full_loss_weight} shows the AP results of Table~\ref{tab:loss_weight}, which compares the various weight parameter $\lambda_{\OP{2D}}$ while keeping $\lambda_{\OP{3D}}=1.0$ in \eqref{eq:loss}. Each point denotes the mean of three trials, and the hatching denotes the standard deviation.
The table shows that it is possible to achieve high performance by using both the loss of the 3D BBox in the radar coordinate system and the loss of the 2D BBox on the image plane.
Also, the figure shows that values of $\lambda_{\OP{2D}} \leq 0.5$ exhibit a larger standard deviation. In other words, there is a general upward trend, reflecting the trade-off between 2D and 3D losses.

\paragraph{Number of BBoxes in Training:}
Table~\ref{tab:full_num_bboxes_training} shows the complete results of Table~\ref{tab:num_bboxes_training}.
From the table, we can see that $\OP{AP}_{50}$ drops significantly for RETR as the number of queries increases, but REXO maintains a high level of accuracy. On the other hand, there is no significant fluctuation in $\OP{AR}$, and RETR and REXO have similar trends. However, REXO has a significantly higher performance.

\paragraph{Dynamic Number of BBoxes in Inference:}
Fig.~\ref{fig:full_comp_n_bbox_inference} shows the AP50 results of Table~\ref{fig:comp_n_bbox_inference}.
The red line shows the results of REXO, and the blue line shows the results of RETR. When the number of BBoxes used in the inference is increased, we can see that RETR performs much worse, while REXO maintains roughly the same performance for all BBoxes. This is because RETR uses queries that require training, and, combined with the results in Table~\ref{tab:full_num_bboxes_training}, a significant deviation from the number of objects in the data will lead to a decrease in performance. On the other hand, REXO always uses random initial BBoxes, so it does not require training and can handle a more flexible number of BBoxes regardless of the number of objects.

\paragraph{Number of Iteration Steps:}
Fig.~\ref{fig:full_comp_diff_steps} shows the AP50 results of Table~\ref{fig:comp_diff_steps}.
The number of steps corresponds to $S$, as explained in Appendix~\ref{sec:detailedDiffusion}. From the figure, we can see that REXO's performance improves as the number of steps $S$ increases.

\paragraph{Impact of Denoising One Parameter:}
Table~\ref{tab:factorization} evaluated two variants: denoising only $c_z$ (center depth) or $d$ (depth extent) for $10$ steps, while fixing the other parameters after $1$ step. 
Both variants maintain strong performance, though jointly denoising all parameters ($10$ steps) yields the best result.

\paragraph{Ground-Level Constraint:}
The effectiveness of prior-constrained 3D BBox diffusion is evaluated in terms of convergence during training. The training curve is shown in Fig.~\ref{fig:full_comp_convergence}. The horizontal axis represents the number of training steps and the vertical axis represents $\OP{AP}$. From this figure, it can be seen that convergence is faster when prior-constrained 3D BBox diffusion is applied than when it is not. Performance also remained high on the validation set when prior-constrained 3D BBox diffusion was applied, suggesting that it improves both fast convergence and generalization performance. 

\begin{figure*}
    \centering
    \includegraphics[width=0.8\linewidth]{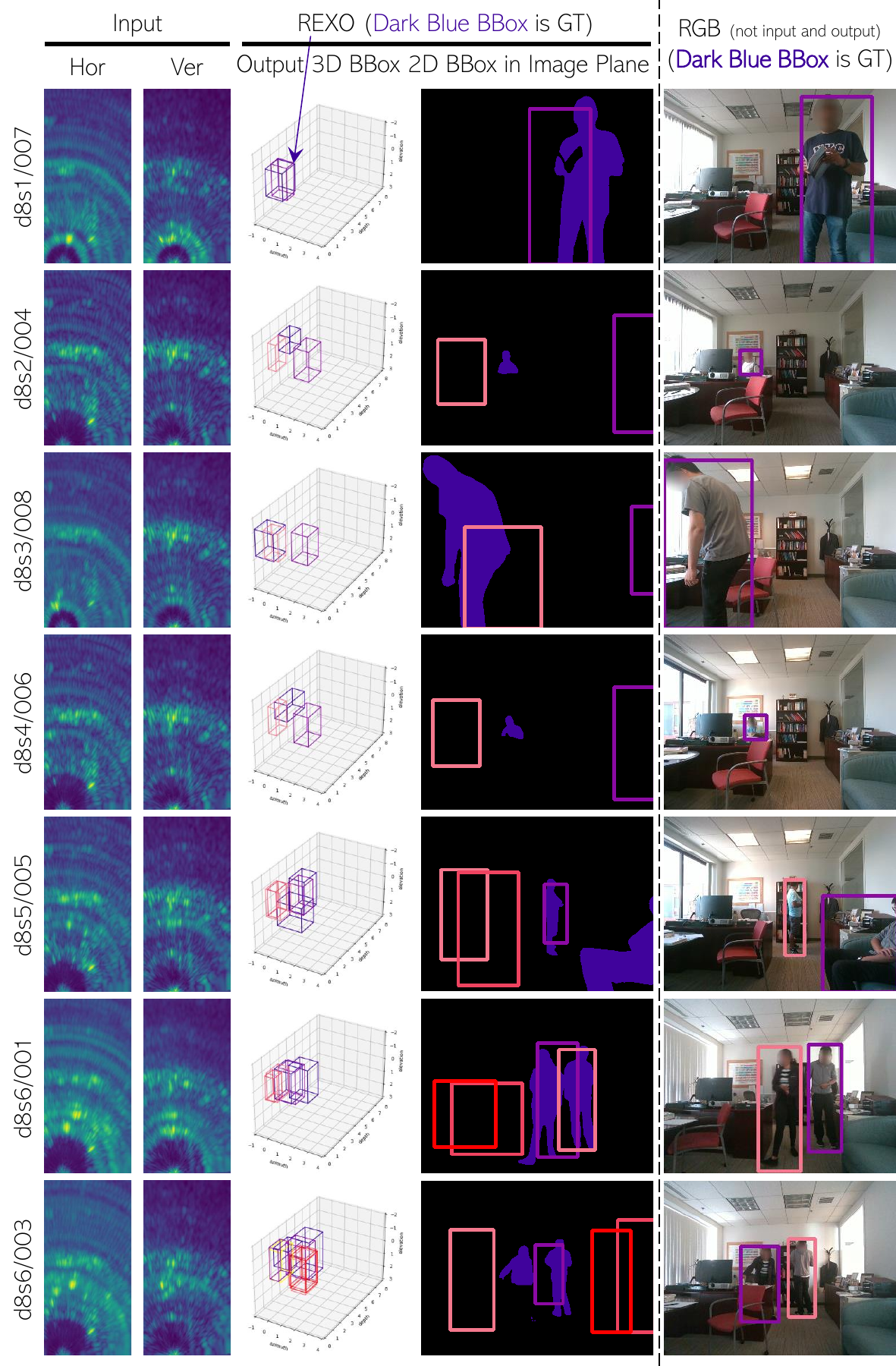}
    \caption{Visualization of failure cases. Each row indicates the segment name used from the P2S2 test dataset.}
    \label{fig:additional_visualization_2}
\end{figure*}

\paragraph{Analysis of Failure Cases:}
We provide failure cases in Fig.~\ref{fig:additional_visualization_2}. 
These are all results of ``Unseen,'' which means the environment that is not included in the training data (d8).
As with d8s1 and d8s3, REXO may sometimes predict inaccurate positions, although less frequently than RETR and RFMask. In addition, there are cases where false negatives occur, such as with d8s2, d8s4, d8s5, and d8s6. In particular, it is thought to be difficult to capture the characteristics of individuals that are far away from the radar, such as with d8s2, because the resolution becomes coarse. In addition, REXO frequently gets false positives such as d8s2 - d8s6, so adjusting the threshold is important.

\clearpage

\end{document}